\RequirePackage{fix-cm}
\documentclass[twocolumn]{svjour3}          

\smartqed  
\usepackage[numbers]{natbib}
\usepackage{indentfirst}
\usepackage{graphicx}
\usepackage[justification=centering]{caption}
\usepackage{subfigure}
\usepackage{amssymb, amsmath}
\usepackage{mathtools}
\usepackage{tabularx}
\usepackage{multirow}
\usepackage{threeparttable}
\usepackage{color}
\usepackage{adjustbox}
\usepackage{url}
\usepackage{soul}
\usepackage{graphicx}
\usepackage{amssymb}
\usepackage{multirow, makecell, tabularx, booktabs}
\usepackage{palatino}

\usepackage{supertabular}
\usepackage{hyperref}
\usepackage{natbib}
\usepackage{geometry}

\setcitestyle{authoryear,round}
\hypersetup{hidelinks,
 colorlinks=true,
 allcolors=black,
 pdfstartview=Fit,
 breaklinks=true
}
\usepackage[section]{placeins}
\begin{document}
	
\title{Video Super-Resolution Based on Deep Learning: A Comprehensive Survey}

\author{Hongying~Liu \and Zhubo~Ruan \and Peng~Zhao \and Chao~Dong \and Fanhua~Shang \and Yuanyuan~Liu \and Linlin~Yang \and Radu Timofte}


\institute{Fanhua~Shang (Corresponding author): \email{fhshang@xidian.edu.cn}\\
   Hongying~Liu and Fanhua~Shang are with the Key Laboratory of Intelligent Perception and Image Understanding of Ministry of Education, School of Artificial Intelligence, Xidian University, China, and Peng Cheng Laboratory, Shenzhen, China.\\
 Zhubo~Ruan, Peng~Zhao, Yuanyuan~Liu and Linlin~Yang are with the Key Laboratory of Intelligent Perception and Image Understanding of Ministry of Education, School of Artificial Intelligence, Xidian University, China.\\
	 Chao~Dong is with the Shenzhen Institutes of Advanced Technology, Chinese Academy of Sciences.\\
	 Radu Timofte is with ETH Zurich, Switzerland and University of Wurzburg, Germany.
}

	\maketitle
	
\begin{abstract}
 Video super-resolution (VSR) is reconstructing high-resolution videos from low resolution ones. Recently, the VSR methods based on deep neural networks have made great progress. However, there is rarely systematical review on these methods. In this survey, we comprehensively investigate 37 state-of-the-art VSR methods based on deep learning. It is well known that the leverage of information contained in video frames is important for video super-resolution. Thus we propose a taxonomy and classify the methods into seven sub-categories according to the ways of utilizing inter-frame information. Moreover, descriptions on the architecture design and implementation details are also included. Finally, we summarize and compare the performance of the representative VSR methods on some benchmark datasets. We also discuss the applications, and some challenges, which need to be further addressed by researchers in the community of VSR. To the best of our knowledge, this work is the first systematic review on VSR tasks, and it is expected to make a contribution to the development of recent studies in this area and potentially deepen our understanding of the VSR techniques based on deep learning.
\end{abstract}

\section{Introduction}	
Super-resolution (SR) aims at recovering a high-resolution (HR) image or multiple images from the corresponding low-resolution (LR) counterparts. It is a classic and challenging problem in computer vision and image processing, and it has extensive real-world applications, such as medical imagery reconstruction \citep{SAINT},  remote sensing~\citep{luo2017video}, and panorama video super-resolution~\citep{360SR,liu2020single}, surveillance systems~\citep{2017Fractional}, and high-definition television~\citep{patti1997superresolu}. With the advent of the 5th generation mobile communication technology, large-sized images or videos can be transmitted within a shorter time. Meanwhile, with the popularity of high-definition (HD) and ultra-high-definition (UHD) display devices, video super-resolution is attracting more attention.
	
Video is one of the most common multimedia in our daily life, and thus super-resolution of low-resolution videos has become very important. In general, image super-resolution methods process a single image at a time, while video super-resolution algorithms deal with multiple successive images/frames at a time so as to utilize relationship within frames to super-resolve the target frame. In a broad sense, video super-resolution (VSR) can be regarded as an extension of image super-resolution and can be processed by image super-resolution algorithms frame by frame. However, the SR performance is not always satisfactory as artifacts and jams may be brought in, which causes unwanted temporal incoherence within frames.
	
\begin{figure*}[!ht]
\centering
\includegraphics[scale=0.69]{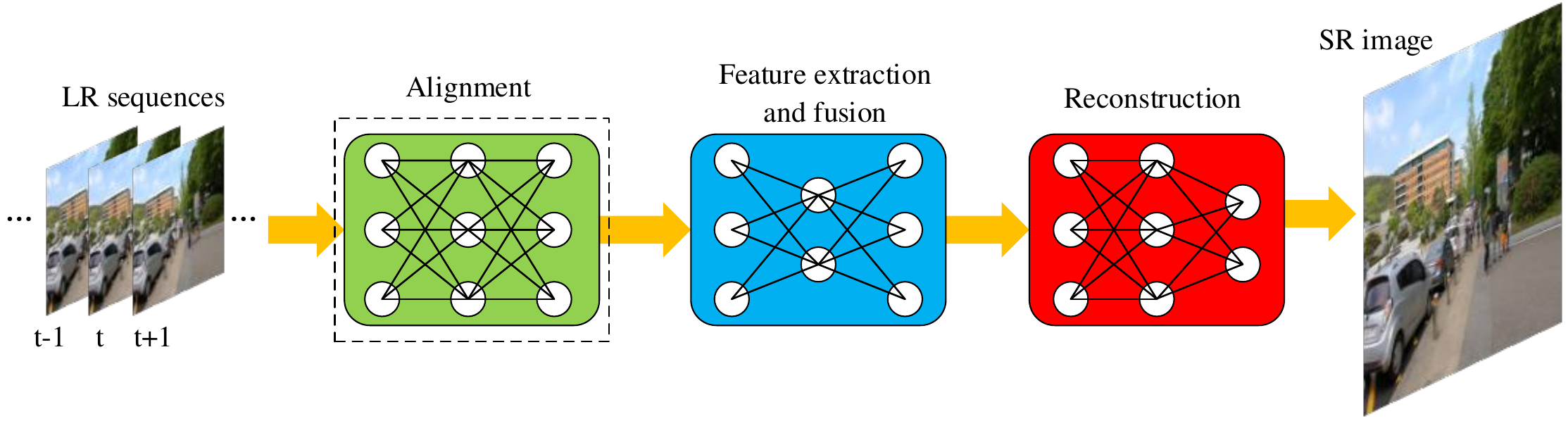}
\caption{The general pipeline of deep learning methods for VSR tasks. Note that the inter-frame alignment module can be either traditional methods or deep CNNs, while both the feature extraction \& fusion module and the upsampling module usually utilize deep CNNs. The dashed line box means that the module is optional.}
\label{VSR_fw}
\end{figure*}

In recent years, many video super-resolution algorithms have been proposed. They mainly fall into two categories: traditional methods and deep learning based methods. For some traditional methods, the motions are simply estimated by affine models as in \citep{TIP1996Extra}. In~\citep{TIP2009NonLocal,TIP2009Wexplit}, they adopt non-local mean and 3D steering kernel regression for video super-resolution, respectively. \citet{liu2013bayesian} proposed a Bayesian approach to simultaneously estimate underlying motion, blur kernel, and noise level for reconstructing high-resolution frames. In~\citep{ma2015handling}, the expectation maximization (EM) method is adopted to estimate the blur kernel, and guide the reconstruction of high-resolution frames. However, these explicit models of high-resolution videos are still inadequate to fit various scenes in videos.

With the great success of deep learning in a variety of areas~\citep{zhang2021application}, super-resolution algorithms based on deep learning are studied extensively. Many video super-resolution methods based on deep neural networks such as convolutional neural network (CNN), generative adversarial network (GAN) and recurrent neural network (RNN) have been proposed. Generally, they employ a large number of both LR and HR video sequences to input the neural network for inter-frame alignment, feature extraction/fusion, and then to produce the high-resolution sequences for the corresponding low-resolution video sequences. The pipeline of most video super-resolution methods mainly includes one alignment module, one feature extraction and fusion module, and one reconstruction module, as shown in Fig.\ \ref{VSR_fw}. Because of the nonlinear learning capability of deep neural networks, the deep learning based methods usually achieve good performance on many public benchmark datasets.

So far, there are few works about the overview on video super-resolution tasks, though many works ~\citep{wang2020deep,singh2020survey,yang2019deep} on the investigation of single image super-resolution have been published. \citet{daithankar2020video} presented a brief review on many frequency-spatial domain methods, while the deep learning methods are rarely mentioned. Unlike the previous work, we provide a comprehensive investigation on deep learning techniques for video super-resolution in recent years. It is well known that the main difference between video super-resolution and image super-resolution lies in the processing of inter-frame information. How to effectively leverage the information from neighboring frames is critical for VSR tasks. We focus on the ways of utilizing inter-frame information for various deep learning based methods.
	
The contributions of this work are mainly summarized as follows. 1) We review recent works and progresses on developing techniques for deep learning based video super-resolution. To the best of our knowledge, this is the first comprehensive survey on deep learning based VSR methods. 2) We propose a taxonomy for deep learning based video super-resolution methods by categorizing their ways of  utilizing inter-frame information and illustrate how the taxonomy can be used to categorize existing methods. 3) We summarize the performance of state-of-the-art methods on some public benchmark datasets, and list the applications of VSR algorithms in various areas. 4) We further discuss some challenges and perspectives for video super-resolution tasks.

The rest of the paper is organized as follows. In Section II, we briefly introduce the background of video super-resolution. Section III shows our taxonomy for recent works. In Sections IV and V, we describe the video super-resolution methods with and without alignment, respectively, according to the taxonomy. In Section VI, the performance of state-of-the-art methods is analyzed quantitatively. In Section VII, we discuss the challenges and prospective trends in video super-resolution. Finally, we conclude this work in Section VIII.

\section{Background}
Video super-resolution stems from image super-resolution, and it aims at restoring high-resolution videos from multiple low-resolution frames. However, the difference between video and image super-resolution techniques is also obvious, that is, the former usually takes advantage of inter-frame information. Besides the RGB color space, the YUV including YCbCr color space is also widely used for VSR. $I_i\!\in\!\mathbb{R}^{H\times W\times3}$ denotes the $i$-th frame in a LR video sequence $I$, and $\hat{I_i}\!\in\!\mathbb{R}^{sH\times sW\times3}$ is the corresponding HR frame, where $s$ is the scale factor, e.g., $s\!=\!2$, 4 or 8. And $\{\hat{I}_j\}_{j=i-N}^{i+N}$ is a set of $2N\!+\!1$ HR frames for the center frame $\hat{I_i}$, where $N$ is the temporal radius. Then the degradation process of HR video sequences can be formulated as follows:
\begin{equation}
{I_i}=\phi(\hat{I}_i,\{\hat{I}_j\}_{j=i-N}^{i+N};\theta_\alpha)
\end{equation}
where $\phi(\cdot;\cdot)$ is the degradation function, and the parameter $\theta_\alpha$ represents various degradation factors such as noise, motion blur and downsampling factors. In most existing works~\citep{liu2013bayesian,ma2015handling, MultiDegra1,MultiDegra4}, the degradation process is expressed as:
\begin{equation}
I_j=DBE_{i\rightarrow j}\hat{I}_i+n_j
\end{equation}
where $D$ and $B$ are the down-sampling and blur operations, $n_j$ denotes image noise, and $E_{i \rightarrow j}$ is the warping operation based on the motion from $\hat{I}_i$ to $\hat{I}_j$.

In practice, it is easy to obtain LR image ${I_j}$, but the degradation factors, which may be quite complex or probably a combination of several factors, are unknown. Different from single image super-resolution (SISR) aiming at solving a single degraded image, VSR needs to deal with degraded video sequences, and recovers the corresponding HR video sequences, which should be as close to the ground truth (GT) videos as possible. Specifically, a VSR algorithm may use similar techniques to SISR for processing a single frame (spatial information), while it has to take relationships among frames (temporal information) into consideration to ensure motion consistency of the video. The super-resolution process, namely the reverse process of Eq.\ (1), can be formulated as follows:
\begin{align}
\tilde{I_i}=\phi^{-1}({I_i},\{{I}_j\}_{j=i-N}^{i+N};\theta_\beta)
\end{align}
where $\tilde{I_i}$ denotes the estimation of the GT (i.e., $\hat{I}_i$), and $\theta_\beta$ is the model parameter.

Like SISR, video quality is mainly evaluated by calculating peak signal-noise ratio (PSNR) and structural similarity index (SSIM). These indexes measure the difference of pixels and similarity of structures between two images, respectively. PSNR of one SR frame is defined as:
\begin{equation}
\textup{PSNR}=10\log_{10}\left(\frac{L^2}{\textup{MSE}}\right)
\end{equation}
where $L$ represents the maximum range of color value, which is usually 255, and the mean squared error (MSE) is defined as:
\begin{equation}
\textup{MSE}=\frac{1}{N}\sum_{i=1}^N(\hat{I}_i-\tilde{I}_i)^2
\end{equation}
where $N$ denotes the total number of pixels in an image or a frame, $\hat{I}$ and $\tilde{I}$ are the ground truth HR frame and the SR recovered frame, respectively. A higher value of PSNR generally means superior quality. In addition, SSIM is defined as:
\begin{equation}
\textup{SSIM}(\hat{I},\tilde{I})=\frac{2u_{\hat{I}}u_{\tilde{I}}+k_1}{u_{\hat{I}}^2+u_{\tilde{I}}^2+k_1}\cdot\frac{2\sigma_{\hat{I}\tilde{I}}+k_2}{\sigma^2_{\hat{I}}+\sigma_{\tilde{I}}^2+k_2}
\end{equation}
where $u_{\hat{I}}$ and $u_{\tilde{I}}$ represent the mean values of the images $\hat{I}$ and $\tilde{I}$, respectively. $k_1$ and $k_2$ are constants, which are used to stabilize the calculation and are usually set to 0.01 and 0.03, respectively.  $\sigma_{\hat{I}}$ and $\sigma_{\tilde{I}}$ denote the standard deviations, and $\sigma_{\hat{I}\tilde{I}}$ denotes the covariance.

\section{Video Super-resolution Methods}
As the videos are a recording of moving visual images and sound, the methods for video super-resolution learn from existing single image super-resolution methods. There are many deep learning based image super-resolution methods such as Super-Resolution using deep Convolutional Neural Networks (SRCNN)~\citep{SRCNN}, Fast Super-Resolution Convolutional Neural Networks (FSRCNN)~\citep{FSRCNN}, VDSR \citep{VDSR}, Efficient Sub-Pixel Convolutional neural Network (ESPCN)~\citep{ESPCN}, Residual Dense Network (RDN)~\citep{RDN}, Residual Channel Attention Network (RCAN)~\citep{RCAN}, ``Zero-Shot" Super-Resolution (ZSSR)~\citep{Zero-Shot} and Super-Resolution using a Generative Adversarial Network (SRGAN)~\citep{SRGAN}. In 2016, based on SRCNN, Kappeler~\citep{VSRnet} presented a video super-resolution method with convolutional neural networks (VSRnet). So far, many video super-resolution algorithms have been proposed. In the following, we summarize the characteristics of the deep learning based methods for video super-resolution in recent years, as shown in Table \ref{VSR_tab}.

\begin{figure*}[!ht]
\centering
\includegraphics[width=1.995\columnwidth]{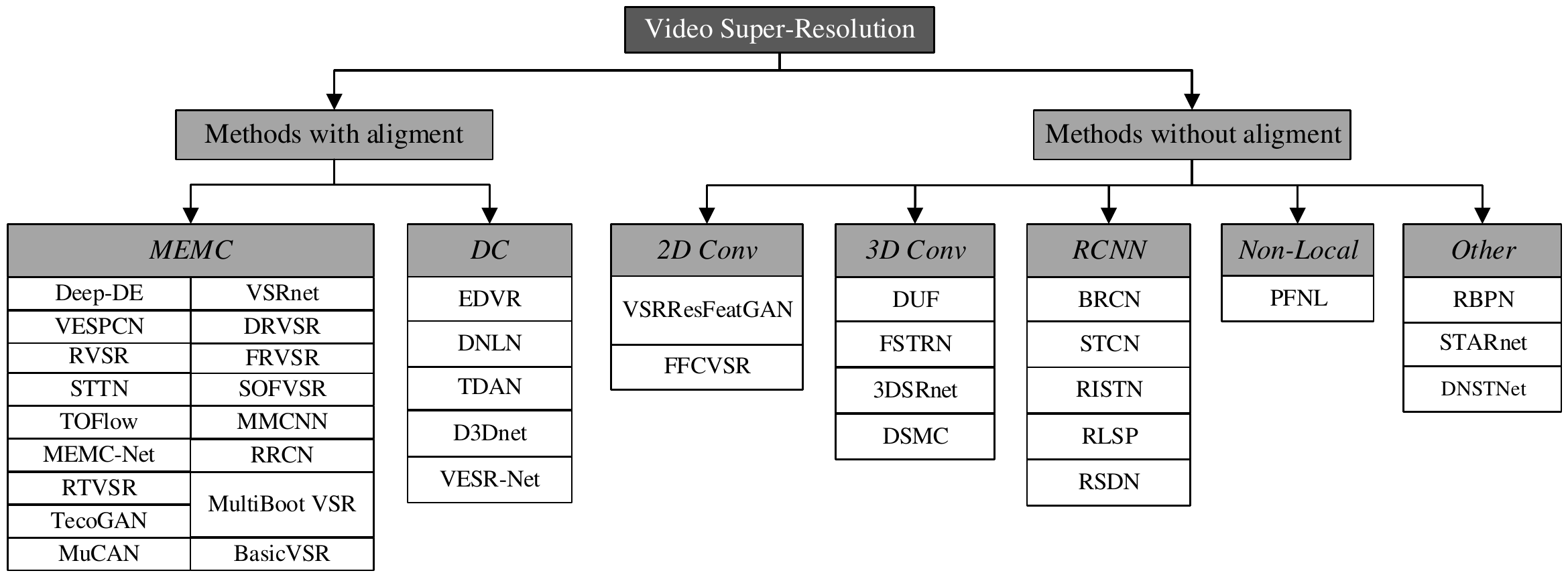}
\caption{A taxonomy for existing state-of-the-art video super-resolution methods. Here, $MEMC$ stands for motion estimation and compensation methods, $DC$ is deformable convolution methods, $3D\ Conv$ is 3D convolution methods, and $RCNN$ denotes recurrent convolutional neural network based methods. The links to these methods are as follows.
\textbf{\emph{MEMC}}: Deep-DE~\citep{Deep-DE},VSRnet~\citep{VSRnet}, VESPCN~\citep{VESPCN}, DRVSR~\citep{DRVSR}, RVSR~\citep{RVSR}, FRVSR~\citep{FRVSR}, STTN~\citep{STTN}, SOFVSR~\citep{SOFVSR}, TOFlow~\citep{TOFlow}, MMCNN~\citep{MMCNN}, MEMC-Net~\citep{MEMC-Net}, RRCN~\citep{8501928}, RTVSR~\citep{RTVSR}, MultiBoot VSR~\citep{MultiBoot},  TecoGAN~\citep{TecoGAN}, MuCAN~\citep{MuCAN}, BasicVSR~\citep{chan2020basicvsr}.
\textbf{\emph{DC}}: EDVR~\citep{EDVR}, DNLN~\citep{DNLN}, TDAN~\citep{TDAN}, D3Dnet~\citep{D3Dnet}, VESR-Net~\citep{chen2020vesr}.
\textbf{\emph{2D Conv}}:VSRResFeatGAN~\citep{VSRResNet}, FFCVSR~\citep{FFCVSR}.
\textbf{\emph{3D Conv}}: DUF~\citep{DUF}, FSTRN~\citep{FSTRN}, 3DSRnet~\citep{3DSRnet}, DSMC~\citep{DSMC2021}.
\textbf{\emph{RCNN}}: BRCN~\citep{NIPS2015_5778, BRCN}, STCN~\citep{STCN2017AAAI}, RISTN~\citep{RISTN}, RLSP~\citep{fuoli2019efficient}, RSDN~\citep{isobe2020video}.
\textbf{\emph{Non-Local}}: PFNL~\citep{PFNL}.
\textbf{\emph{Other}}: RBPN~\citep{RBPN}, STARnet~\citep{STARnet}, DNSTNet~\citep{sun2020video}.
}
\label{classification}
\end{figure*}
	
Several recent studies such as \citep{EDVR,DUF,TDAN} on video super-resolution tasks have indicated that the utilization of the information contained in frames greatly influences performance. The proper and adequate usage of such information can enhance the results of video super-resolution. Therefore, we build a taxonomy for existing video super-resolution methods according to their ways of the utilization of inter-frame information, as shown in Fig.\ \ref{classification}.

\begin{table*}[!h]
\caption{Existing video super-resolution methods based on deep learning and their key strategies such as loss functions (see their source papers for the details of the loss functions). Here, $MEMC$ denotes motion estimation and motion compensation, $DC$ is deformable convolution, $3D\ Conv$ is 3D convolution, and $RCNN$ denotes recurrent convolutional neural networks.}
    \centering
    \renewcommand\arraystretch{1.39}
    \resizebox{\textwidth}{!}{
    \begin{tabular}{|c|c|m{6cm}|c|m{4.6cm}|c|c|}
    \hline
        \textbf{Method} & \textbf{Year} & \textbf{Synonym} & \textbf{Type} & \textbf{Loss function } & \!\textbf{Align}\!\\
        \hline
        Deep-DE~\citep{Deep-DE} & ICCV 2015 & Deep Draft-Ensemble Learning & \multirow{24}{*}{MEMC} &$\ell_1$-norm loss with total variation regularization & $\checkmark$ \\
         \cline{1-3} \cline{5-6}
        VSRnet~\citep{VSRnet} & TCI 2016 & Video Super-Resolution with convolutional neural Networks & ~ & Mean Square Error (MSE) loss & $\checkmark$ \\
        \cline{1-3} \cline{5-6}
        VESPCN~\citep{VESPCN} & CVPR 2017 & Video Efficient Sub-pixel Convolutional Network & ~ & MSE loss and Motion Compensation (MC) loss & $\checkmark$ \\
        \cline{1-3} \cline{5-6}
        DRVSR~\citep{DRVSR} & ICCV 2017 & Detail-Revealing deep Video Super-Resolution & ~ & MSE loss and MC loss & $\checkmark$ \\ \cline{1-3} \cline{5-6}
        RVSR~\citep{RVSR} & ICCV 2017 & Robust Video Super-Resolution & ~ & Spatial alignment loss and spatio-temporal adaptive loss & $\checkmark$ \\ \cline{1-3} \cline{5-6}
        FRVSR~\citep{FRVSR} & CVPR 2018 & Frame-Recurrent Video Super-Resolution & ~ & MSE loss and MC loss & $\checkmark$ \\ \cline{1-3} \cline{5-6}
        STTN~\citep{STTN} & ECCV 2018 & Spatio-Temporal Transformer Network & ~ & MSE loss and MC loss & $\checkmark$ \\ \cline{1-3} \cline{5-6}
        SOFVSR~\citep{SOFVSR} & ACCV 2018 & Super-resolution Optical Flow for Video Super-Resolution & ~ & MSE loss and MC loss & $\checkmark$ \\ \cline{1-3} \cline{5-6}
        TOFlow~\citep{TOFlow} & IJCV 2019 & video enhancement with Task-Oriented Flow & ~ & $\ell_1$-norm loss & $\checkmark$ \\ \cline{1-3} \cline{5-6}
        MMCNN~\citep{MMCNN} & TIP 2019 & Multi-Memory Convolutional Neural Network & ~ & MSE loss and MC loss & $\checkmark$ \\ \cline{1-3} \cline{5-6}
        MEMC-Net~\citep{MEMC-Net} & TPAMI 2019 & Motion Estimation and Motion Compensation Network & ~ & Charbonnier (Cb) loss & $\checkmark$  \\ \cline{1-3} \cline{5-6}
        RRCN~\citep{8501928} & TIP 2019 & Residual Recurrent Convolutional Network & ~ & MSE loss & $\checkmark$ \\ \cline{1-3} \cline{5-6}
        RTVSR~\citep{RTVSR} & Neurocomp. 2019 & Real-Time Video Super-Resolution & ~ & MSE loss & $\checkmark$ \\ \cline{1-3} \cline{5-6}
        \!\!MultiBoot VSR~\citep{MultiBoot}\!\! & CVPRW 2019 & Multi-stage multi-reference Bootstrapping for Video Super-Resolution & ~ & Huber loss & $\checkmark$ \\  \cline{1-3} \cline{5-6}
        MuCAN~\citep{MuCAN} & ECCV 2020 & Multi-Correspondence Aggregation Network for Video Super-Resolution & ~ & Edge-aware loss & $\checkmark$ \\ \cline{1-3} \cline{5-6}
        TecoGAN~\citep{TecoGAN} & ACMTOG 2020 & Temporally coherent GAN & ~ & MSE loss and ping-pong loss etc. & $\checkmark$ \\ \cline{1-3} \cline{5-6}
        BasicVSR~\citep{chan2020basicvsr} &CVPR 2021 & search for essential components in Video Super-Resolution and beyond & ~ & Cb loss & $\checkmark$ \\ \hline
        EDVR~\citep{EDVR} & CVPRW 2019 & Enhanced Deformable convolutional networks for Video Restoration & \multirow{11}{*}{DC} &  Cb loss & $\checkmark$ \\
        \cline{1-3} \cline{5-6}
        DNLN~\citep{DNLN} & ACCESS 2019 & Deformable Non-Local Network for Video Super-Resolution & ~ & $\ell_1$-norm loss & $\checkmark$  \\ \cline{1-3} \cline{5-6}
        TDAN~\citep{TDAN} & CVPR 2020 & Temporally-Deformable Alignment Network for Video Super-Resolution & ~ & $\ell_1$-norm loss & $\checkmark$ \\ \cline{1-3} \cline{5-6}
        D3Dnet~\citep{D3Dnet} & SPL 2020 & Deformable 3D Convolution for Video Super-Resolution & ~ & MSE loss & $\checkmark$ \\ \cline{1-3} \cline{5-6}
        VESR-Net~\citep{chen2020vesr} & ArXiv 2020 & Video Enhancement and Super-Resolution Network & ~ & $\ell_1$-norm loss & $\checkmark$ \\ \hline
        \!VSRResFeatGAN~\citep{VSRResNet}\! &TIP 2019 & Video Super-Resolution with Residual Networks & \multirow{3}{*}{2D Conv} & Adversarial loss; content loss; and  perceptual loss & $\times$ \\ \cline{1-3} \cline{5-6}
        FFCVSR~\citep{FFCVSR} & AAAI 2019 & Frame and Feature-Context Video Super-Resolution & ~ & MSE loss & $\times$ \\ \hline
        DUF~\citep{DUF} &  CVPR 2018 & video super-resolution network using Dynamic Upsampling Filters &  \multirow{6}{*}{3D Conv} & Huber loss & $\times$  \\ \cline{1-3} \cline{5-6}
        FSTRN~\citep{FSTRN} & CVPR 2019 & Fast Spatio-Temporal Residual Network for Video Super-Resolution & ~ & Cb loss & $\times$  \\ \cline{1-3} \cline{5-6}
        3DSRnet~\citep{3DSRnet} & ICIP 2019 & 3D Super-Resolution Network & ~ & MSE loss & $\times$ \\ \cline{1-3} \cline{5-6}
        DSMC~\citep{DSMC2021} & AAAI 2021 & Dual Subnet and Multi-stage Communicated upsampling & ~ & Cb loss; perceptual loss; the dual loss & $\times$ \\ \hline
        \!\!\!BRCN~\citep{NIPS2015_5778, BRCN}\!\!\! & \!NIPS 2015/2018\! & video super-resolution via Bidirectional Recurrent Convolutional Networks & ~ & MSE loss & $\times$  \\ \cline{1-3} \cline{5-6}
        STCN~\citep{STCN2017AAAI} & AAAI 2017 & Spatio-Temporal Convolutional Network for Video Super-Resolution & \multirow{5}{*}{RCNN} & MSE loss & $\times$ \\ \cline{1-3} \cline{5-6}
        RISTN~\citep{RISTN} & AAAI 2019 & Residual Invertible Spatio-Temporal Network for Video Super-Resolution & ~ & MSE loss & $\times$
        \\ \cline{1-3} \cline{5-6}
        RLSP~\citep{fuoli2019efficient} & \!ICCVW 2019\! & video super-resolution through Recurrent Latent Space Propagation & ~ & MSE loss & $\times$  \\ \cline{1-3} \cline{5-6}
        RSDN~\citep{isobe2020video} & \!ECCV 2020\! & video super-resolution with Recurrent Structure-Detail Network & ~ & Cb loss & $\times$
        \\ \hline
        PFNL~\citep{PFNL} & ICCV 2019 & Progressive Fusion network via exploiting Non-Local spatio-temporal correlations & \multirow{1}{*}{\!Non-Local\!} & Cb loss & $\times$ \\ \hline
         RBPN~\citep{RBPN} & CVPR 2019 & Recurrent Back-Projection Network & ~ & $\ell_1$-norm loss &$\times$\\ \cline{1-3} \cline{5-6}
        STARnet~\citep{STARnet} & CVPR 2020 & Space-Time-Aware multi-Resolution network & ~ & Three losses & $\times$ \\ \cline{1-3} \cline{5-6}
         DNSTNet~\citep{sun2020video} & Neurocomp. 2020 & video super-resolution via Dense Non-local Spatial-Temporal convolutional Network & \multirow{1}{*}{\!Other \!} & $\ell_1$-norm loss & $\times$ \\ \cline{1-3} \cline{5-6}
        \hline
    \end{tabular}}
    \label{VSR_tab}
\end{table*}

As shown in Fig.\ \ref{classification} and Table \ref{VSR_tab}, we categorize the existing methods into two main categorises: methods with alignment and methods without alignment, according to whether the video frames are explicitly aligned. We will present the methods in detail in the following sections.

Since all the methods are classified according to whether the frames are explicitly aligned and what techniques they are mainly used for alignment, other modules which they utilized for feature extraction, fusion, and reconstruction are ignored. These modules may be employed by multiple methods simultaneously. Therefore, some of the methods in our study are coupled.
BasicVSR in the MEMC methods from the category of the methods with alignment adopts a typical bidirectional recurrent convolutional neural network (RCNN) as backbone. While the RCNN-based methods (e.g. BRCN, STCN, and RISTN), which are in the methods without alignment, mainly use RCNN to learn features. Similarly, VESR-Net in the DC category also uses a non-local block for feature learning as that of PFNL in the non-local category. Moreover, DSMC in the category of 3D convolution also utilizes a non-local block for global correlation computation. The category of 'other' includes the methods which adopt optical flow but without frame alignment, e.g., RBPN, and STARnet. Finally, the learned offsets by deformable convolution share similar patterns as those from the optical flow-based methods, and the deformable and flow-based alignments are strongly correlated. This was indicated in the work \citep{chan2021understanding}.

Moreover, we have observed several trends in these recently proposed methods.

1) The diversification of methods. In the early years (2015-2017), most of the methods use frame alignment for VSR. Then since 2018, many different methods, especially which are the methods without alignment, have emerged, e.g., FFCVSR, DUF, RISTN, and PLNL. Some studies also indicate that both the methods with alignment and those without alignment can obtain sound performance.

2) The expansion of receptive field in methods. Earlier methods such as EDVR and RBPN mainly utilize certain numbers of input frames in sliding-window, while the subsequent methods resort to longer sequences. For example, BasicVSR employs bidirectional RCNN, by which the features are propagated forward and backward independently. Moreover, the non-local subnetwork, such as in the PFNL method, aims to compute the correlations between all possible pixels within and across frames. These indicate that the methods tend to capture longer-range dependencies in the video sequences, and they expand the receptive field in the network from local to global.

3) In the MEMC methods such as FRVSR, STTN, SOFVSR, TecoGAN, and MuCAN, most of them adopt deep learning techniques for estimation the optical flow, since the deep learning may have adaptive ability for various data than the conventional methods. 4) The practicality of methods. As the requirements for super-resolving of higher quality videos develop, the recently proposed methods also become more practicable. The test videos evolve from Vid4 and UVGD, to REDS. All the discussions indicate that we will mainly focus on the methods for the videos with more complex motions and scene changes.

\section{Methods with Alignment}
The methods with alignment make neighboring frames explicitly align with the target frame by using extracted motion information before subsequent reconstruction. These methods mainly use motion estimation and motion compensation (MEMC) or deformable convolution, which are two common techniques for aligning frames. Next we will introduce state-of-the-art methods based on each of the techniques in detail.

\begin{figure*}[!th]
\center
\subfigure[A target frame]{\includegraphics[width=0.68\columnwidth]{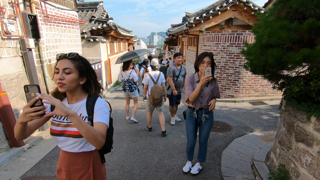}}\qquad\;\;\subfigure[Its neighboring frame]{\includegraphics[width=0.68\columnwidth]{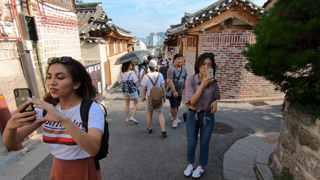}}\\
\subfigure[The compensated image]{\includegraphics[width=0.68\columnwidth]{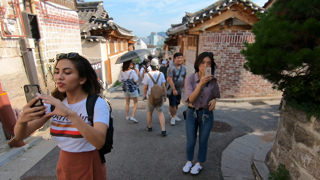}}\qquad\;\;\subfigure[The estimated optical flow image]{\includegraphics[width=0.68\columnwidth]{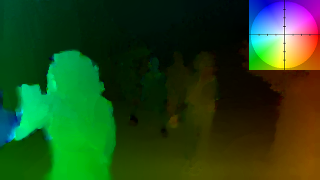}}
\caption{An example of motion estimation and compensation. Note that the small rightmost image is the legend of (d). Different colors represent different directions of motion and the intensity of the color is the range of motion.}
\label{memc}
\end{figure*}

\subsection{Motion Estimation and Compensation Methods}
In the methods with alignment for video super-resolution, most of them apply the motion estimation and motion compensation techniques. Specifically, the purpose of motion estimation is to extract inter-frame motion information, while motion compensation is used to perform the warping operation between frames according to inter-frame motion information and to make one frame align with another frame. A majority of the motion estimation techniques are performed by the optical flow method \citep{FlowNet}. This method tries to calculate the motion between two neighboring frames through their correlations and variations in the temporal domain. The motion estimation methods can be divided into two categories: traditional methods (e.g., \citep{Iterative1981} and \citep{TV}) and deep learning methods such as FlowNet \citep{FlowNet}, FlowNet 2.0 \citep{FlowNet2} and SpyNet \citep{Ranjan_2017_CVPR}.

In general, an optical flow method takes two frames (e.g., $I_i$ and $I_j$) as inputs. One is the target frame and the other is the neighboring frame. Then the method computes a vector field of optical flow $F_{i\rightarrow{j}}$ from the frame $I_i$ to $I_j$ by the following formula:
\begin{equation}
F_{i\rightarrow{j}}(h_{i\rightarrow{j}},v_{i\rightarrow{j}})=ME(I_i, I_j; \theta_{ME})
\end{equation}
where $h_{i\rightarrow{j}}$ and $v_{i\rightarrow{j}}$ is the horizontal and vertical components of $F_{i\rightarrow{j}}$, $ME(\cdot)$ is a function used to compute optical flow, and $\theta_{ME}$ is its parameter.

The motion compensation is used to perform image transformation between images in terms of motion information to make neighboring frames align with the target frame. In general, a compensated frame $I_j'$ is expressed as:
\begin{equation}
I_j'=MC(I_i,F_{i\rightarrow{j}};\theta_{MC})
\end{equation}
where $MC(\cdot)$ is a motion compensation function, $I_i$, $F_{i\rightarrow{j}}$ and $\theta_{MC}$ are the neighboring frame, optical flow and the parameter. MC can be achieved by some methods such as bilinear interpolation and spatial transformer network (STN) \citep{STN}. An example of motion estimation and motion compensation is shown in Fig.\ \ref{memc}.

Both the ME and MC processes can be conducted by a deep learning method or traditional one (non-deep learning). According to the technique that applies to ME or MC is traditional or deep learning, we further divide the MEMC methods into two subcategories. If any of the processes in ME or MC utilizes a deep neural network, then the method falls into the deep learning category, otherwise the method belongs to the traditional one. Therefore, the traditional methods in the MEMC methods comprise of the following three ones: Deep-DE~\citep{Deep-DE}, VSRNet~\citep{VSRnet}, and RRCN~\citep{8501928}. The other MEMC methods are included in the deep learning subcategory. Below we depict some representative methods in detail.

\begin{figure}[!htbp]
\centering
\includegraphics[width=0.839\columnwidth]{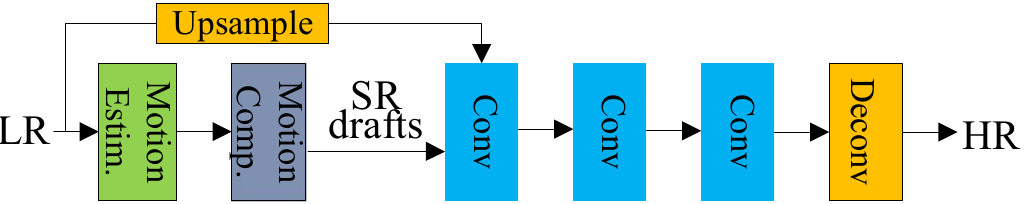}
\caption{The architecture of Deep-DE \citep{Deep-DE}. Here Motion Estim.\ is a motion estimation block, Motion Comp.\ is a motion compensation block, Conv is a convolutional layer and Deconv is a deconvolutional layer.}
\label{figs10001}
\end{figure}

\subsubsection{Deep-DE}
The deep draft-ensemble learning method (Deep-DE)\protect\footnote{Code: http://www.cse.cuhk.edu.hk/leojia/projects/DeepSR/} \citep{Deep-DE} has two phases, as shown in Fig.\ \ref{figs10001}. It first generates a series of SR drafts by adjusting the TV-$\ell_1$ flow \citep{10.1007/978-3-540-24673-2_3,STCN2017AAAI} and the motion detail preserving (MDP) \citep{6104059}. Then both the SR drafts and the bicubic-interpolated LR target frame are fed into a CNN for feature extraction, fusion and super-resolution.

The CNN in Deep-DE consists of four convolutional layers: the first three layers are general convolutional layers, and the last layer is a deconvolution layer. Their kernel sizes are 11$\times$11, 1$\times$1, 3$\times$3 and 25$\times$25, respectively, and the numbers of channels are 256, 512, 1 and 1.
\begin{figure}[!htbp]
\centering
\includegraphics[width=0.819\columnwidth]{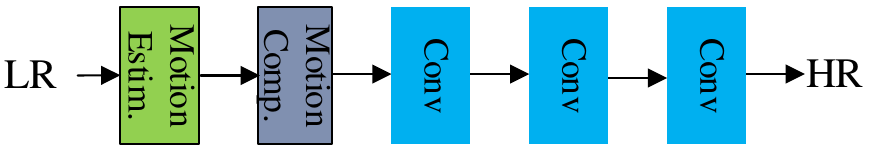}
\caption{The network architecture of VSRnet \citep{VSRnet}.}
\label{figs1001}
\end{figure}

\subsubsection{VSRnet} 
VSRnet\protect\footnote{Code: https://superresolution.tf.fau.de/} \citep{VSRnet} is based on the image super-resolution method, SRCNN \citep{SRCNN}, and its network architecture is shown in Fig.\ \ref{figs1001}. VSRnet mainly consists of motion estimation and compensation modules, and three convolutional layers, and each convolutional layer is followed by a rectified linear unit (ReLU) except for the last one. The main difference between VSRnet and SRCNN is the number of input frames. That is, SRCNN takes a single frame as input, while VSRnet uses multiple successive frames, which are compensated frames. The motion information between frames is computed by the Druleas algorithm \citep{TV}. In addition, VSRnet proposes a filter symmetry enforcement (FSE) mechanism and an adaptive motion compensation mechanism, which are separately used to accelerate training and reduce the impact of unreliable compensated frames, and thus can improve video super-resolution performance.

\begin{figure}[!htbp]
\centering
\includegraphics[width=0.869\columnwidth]{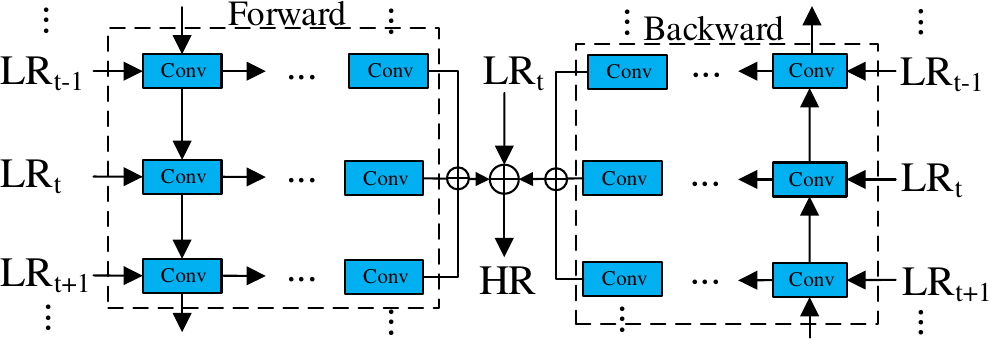}
\caption{The network architecture of RRCN \citep{8501928}.}
\label{figs1026}
\end{figure}

\subsubsection{RRCN}
The residual recurrent convolutional network (RRCN) \citep{8501928}, as shown in Fig.\ \ref{figs1026}, is a bidirectional recurrent neural network, which learns a residual image. RRCN proposes an unsynchronized full recurrent convolutional network, where unsynchronization refers to the input of multiple consecutive video frames, and only the middle one is super-resolved.

RRCN uses the combined local-global with total variable (GLG-TV) method \citep{TV} to perform motion estimation and compensation for the target frame and its adjacent frames. The compensated frames are used as input to the network. The forward convolution and recurrent convolution are conducted in the forward network and the backward network, respectively, and their outputs are summed up. Finally, the result is obtained by adding the target frame to the input. In order to further improve the performance, RRCN also uses the self-ensemble strategy and combines it with the output of the single image super-resolution method, EDSR+ \citep{EDSR}, to obtain two models named RRCN+ and RRCN++, respectively.


\begin{figure}[!htbp]
\centering
\includegraphics[width=0.835\columnwidth]{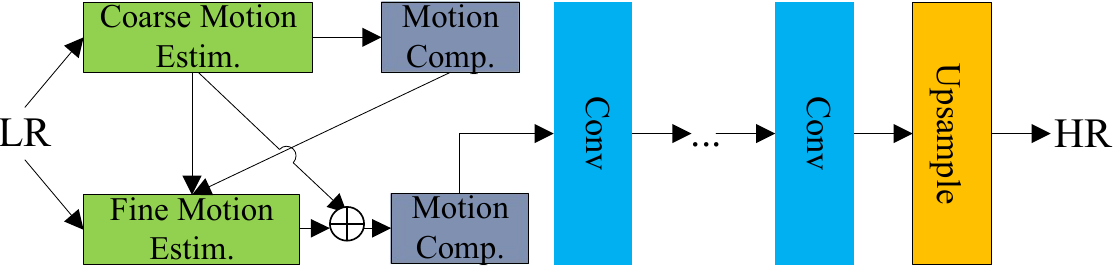}
\caption{The network architecture of VESPCN \citep{VESPCN}. Here $\oplus$ denotes element-wise sum.}
\label{figs1002}
\end{figure}	

\subsubsection{VESPCN}
The video efficient sub-pixel convolutional network (VESPCN) \citep{VESPCN} proposes a spatial motion compensation transformer (MCT) module for motion estimation and compensation. Then the compensated frames are fed into a series of convolutional layers for feature extraction and fusion, as shown in Fig.\ \ref{figs1002}. Finally, the super-resolution results are obtained through a sub-pixel convolutional layer for upsampling.

The MCT module adopts CNNs to extract motion information and perform motion compensation. MCT uses a coarse-to-fine approach to compute the optical flow for image sequences. Firstly, in the coarse estimation stage, the network takes two consecutive frames (i.e., the target frame and a neighboring frame) as inputs. The coarse network consists of 5 convolutional layers and a sub-pixel convolutional layer. And it first performs the $\times$2 downsampling operation two times and then performs the $\times$4 upsampling operation by a sub-pixel convolutional layer to get coarse optical flow estimation results. Secondly, the neighboring frame is warped according to the optical flow. In the fine estimation stage, the target frame, neighboring frame, optical flow computed in the coarse stage and the warped neighboring frame are the input of the fine network, whose architecture is similar to the coarse network. It first conducts $\times$2 downsampling and then perform $\times$2 upsampling at the end of the network to attain the fine optical flow. Together with the coarse optical flow, the fine optical flow is used to obtain the final estimation result. Finally, the neighboring frame is warped again by the final optical flow to make the warped frame align with the target frame.

\begin{figure}[!htbp]
\centering
\includegraphics[width=0.996\columnwidth]{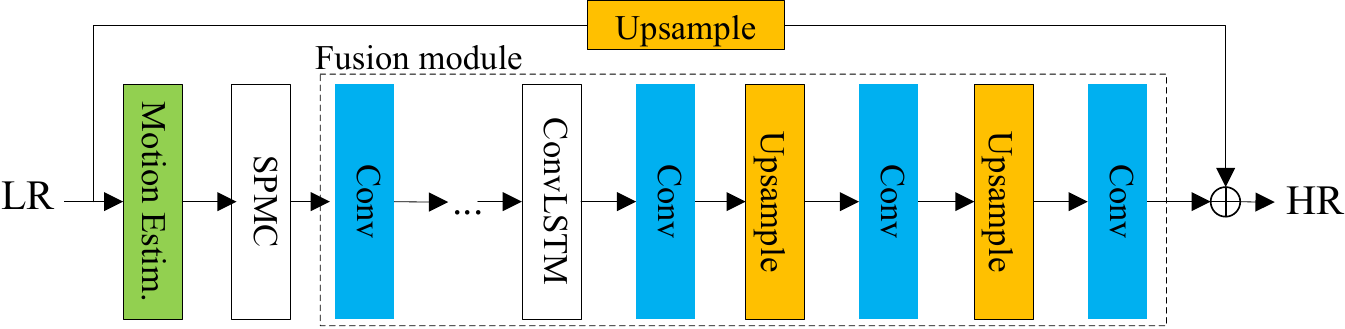}
\caption{The network architecture of DRVSR \citep{DRVSR}. Here SPMC denotes a sub-pixel motion compensation layer, ConvLSTM is the convolutional LSTM \citep{convLSTM}.}
\label{figs1003}
\end{figure}

\subsubsection{DRVSR}  
The detail-revealing deep video super-resolution (DRVSR)\protect\footnote{Code: https://github.com/jiangsutx/SPMC\_VideoSR} \citep{DRVSR} method proposes a sub-pixel motion compensation layer (SPMC) that can perform the up-sampling and motion compensation operations simultaneously for neighboring input frames according to the estimated optical flow information. The network architecture of DRVSR is illustrated in Fig.\ \ref{figs1003}.

DRVSR consists of three main modules: a motion estimation module, a motion compensation module using the SPMC layer, and a fusion module. The motion estimation module is implemented by the motion compensation transformer (MCT) network \citep{VESPCN}. The SPMC layer consists of two sub-modules, namely grid generator and sampler. The grid generator first transforms the coordinates in the LR space into the coordinates in the HR space according to the optical flow, and then the sampler performs the interpolation operation in the HR space. In the fusion module, it applies the convolution with stride 2 to perform down-sampling and then conducts the deconvolution for up-sampling to obtain the HR residual image of the target frame. Together with the upsampled LR target frame, this residual image yields the final result. DRVSR also adopts the ConvLSTM module \citep{convLSTM} to handle spatio-temporal information.

\begin{figure}[!htbp]
\centering
\includegraphics[width=0.727\columnwidth]{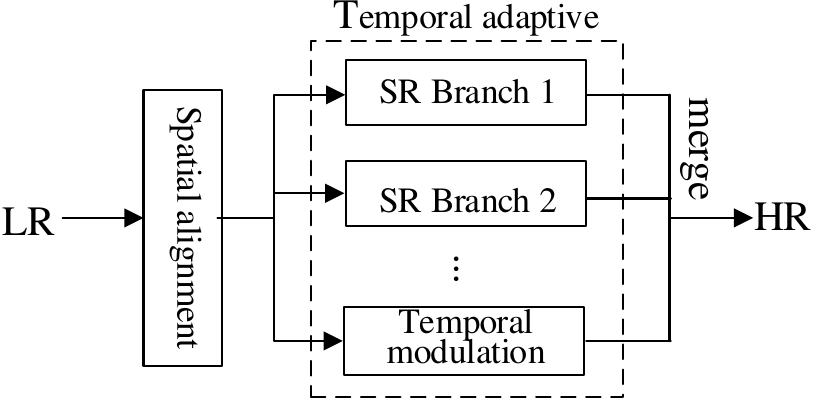}
\caption{The network architecture of RVSR \citep{RVSR}, where SR denotes Super-Resolution.}
\label{figs1004}
\end{figure}

\subsubsection{RVSR}  
Robust video super-resolution (RVSR) \citep{RVSR} proposes a spatial alignment module to attain great alignment performance and a temporal adaptive module to adaptively determine the optimal scale of temporal dependency. And its architecture is shown in Fig.\ \ref{figs1004}.

The spatial alignment module is responsible for the alignment of the multi-frames so that the neighboring frames are aligned with the target frame. It first estimates the transformation parameters between the neighboring frame and the target frame through a localization net, and then makes the neighboring frame align with the target frame through a spatial transformation layer \citep{STN} based on the obtained parameters. The localization net consists of two convolutional layers, each of which is followed by a max-pooling layer, and two fully connected layers. The temporal adaptive module is composed of multiple branches of SR subnetwork and a temporal modulation. Each subnetwork is responsible for handling a temporal scale (i.e., the number of input frames), and outputting the corresponding super-resolution result. Then the super-resolution result of each subnetwork is allocated a weight through the temporal modulation. The final super-resolution result is the weight sum of the super-resolution result of each branch and its weight. The number of the input frames of the temporal modulation module is identical to the maximum number of input frames in the super-resolution network, and the network structure of the temporal modulation module is the same as that of the super-resolution network, and both of them are based on the structure of ESPCN \citep{ESPCN}.

\begin{figure}[!htbp]
\centering
\includegraphics[width=0.725\columnwidth]{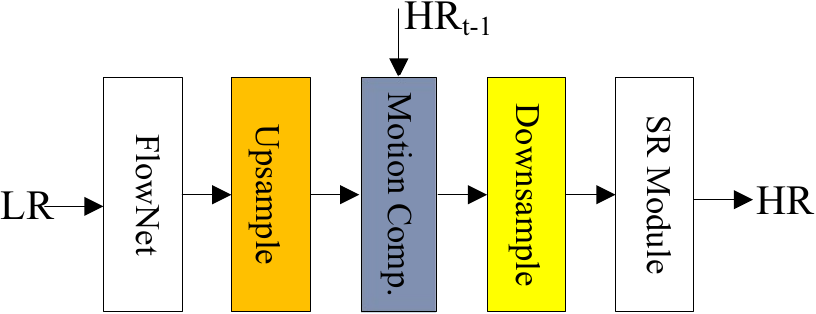}
\caption{The network architecture of FRVSR \citep{FRVSR}. Here FlowNet is an optical flow estimation module, and SR Module is a super-resolution module.}
\label{figs1005}
\end{figure}

\subsubsection{FRVSR}  
Frame recurrent video super-resolution (FRVSR)\protect\footnote{Code: https://github.com/msmsajjadi/FRVSR} \citep{FRVSR} mainly proposes to use the previously inferred HR estimate to super-resolve the subsequent frame for producing temporally consistent results and reducing computational cost. The architecture of FRVSR is illustrated in Fig.\ \ref{figs1005}.

The detailed implementation adopts an optical estimation network to compute the optical flow from the previous frame to the target frame. Then the LR optical flow is upsampled to the same size with the HR video by bilinear interpolation. The HR variant of the previous frame is warped by the upsampled LR optical flow, and then the warped HR frame is downsampled by space-to-depth transformation to get the LR version. Finally, the LR variant of the warped HR frame and the target frame are fed into the subsequent super-resolution network to attain the result for the target frame. In FRVSR, the optical flow network consists of 14 convolutional layers, 3 pooling layers and 3 bilinear upsampling layers. Each convolutional layer is followed by a LeakyReLU activation function, except for the last convolutional layer. The super-resolution network consists of 2 convolutional layers, 2 deconvolution layers with $\times$2 and 10 residual blocks, where each residual block consists of 2 convolutional layers and a ReLU activation function.


\begin{figure}[!htbp]
\centering
\includegraphics[width=0.63\columnwidth]{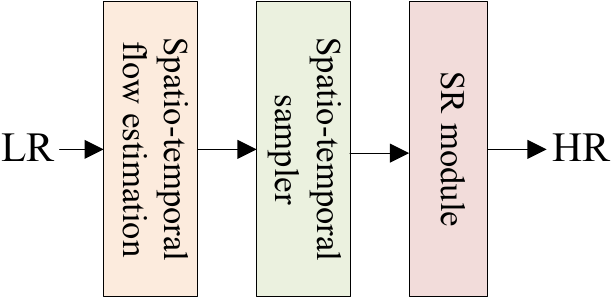}
\caption{The network architecture of STTN \citep{STTN}.}
\label{figs1006}
\end{figure}

\subsubsection{STTN} 
Spatio-temporal transformer network (STTN)  \citep{STTN} proposes a spatio-temporal transformer module, which is used to address the problem that previous optical flow methods only process a pair of video frames, which may cause inaccurate estimation when occlusion and luminance variation exist in videos. The proposed module can handle multiple frames at a time. The architecture of STTN is illustrated in Fig.\ \ref{figs1006}.

STTN consists of three major modules: a spatio-temporal flow estimation module, a spatio-temporal sampler module, and a super-resolution module. The first module is a U-style network, similar to U-Net \citep{u-net}, consisting of 12 convolutional layers and two up-sampling layers. It first performs $\times 4$ downsampling, and then $\times 4$ up-sampling to restore the size of the input frames. This module is responsible for optical flow estimation of the consecutive input frames including the target frame and multiple neighboring frames, and the final output is a 3-channel spatio-temporal flow that expresses the spatial and temporal changes between frames. The spatio-temporal sampler module is actually a trilinear interpolation method, which is responsible for performing warp operation for current multiple neighboring frames and obtaining the aligned video frames according to the spatio-temporal flow obtained by the spatio-temporal flow module. For video super-resolution, the aligned frames can then be fed into the super-resolution (SR) module for feature fusion and super-resolution of the target frame.
	
\begin{figure}[!htbp]
\centering
\includegraphics[width=0.6269\columnwidth]{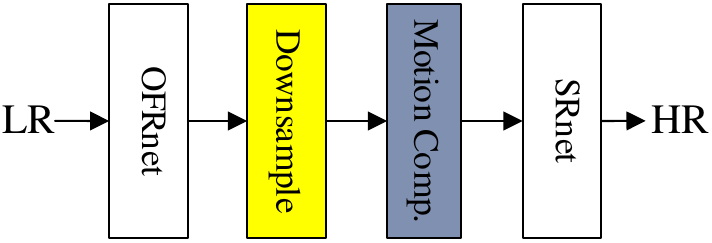}
\caption{The network architecture of SOFVSR \citep{SOFVSR}. Here, OFRnet is an optical flow network, and SRnet is a super-resolution module.}
\label{figs1007}
\end{figure}

\subsubsection{SOFVSR}  
Super-resolution optical flow for video super-resolution tasks (SOFVSR)\protect\footnote{Code: https://github.com/LongguangWang/SOF-VSR} \citep{SOFVSR} is proposed to super-resolve LR estimated optical flow for attaining great SR performance, and its architecture is shown in Fig.\ \ref{figs1007}.

The optical flow between frames is estimated by a coarse-to-fine approach including the optical flow reconstruction network (OFRnet), which finally yields a high-resolution optical flow. Then the HR optical flow is converted to the LR optical flow by a space-to-depth transformation. The neighboring frames are warped by the LR optical flow to make the neighboring frames align with the target frame. Then the super-resolution network (SRnet) takes the target frame and warped frames as inputs to obtain the final super-resolution result. SRnet consists of two convolutional layers, five residual dense blocks and a sub-pixel convolutional layer.

\begin{figure}[!htbp]
\centering
\includegraphics[width=0.735\columnwidth]{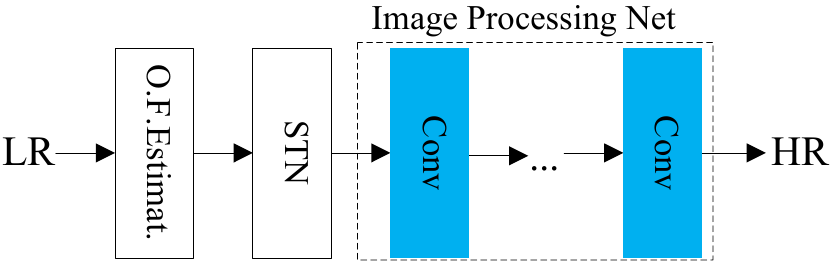}
\caption{The network architecture of TOFlow \citep{TOFlow}. Here O.F.Estimat. is the optical flow estimation, STN is a spatial transformer network.}
\label{figs1009}
\end{figure}

\subsubsection{TOFlow}  
The architecture of the task-oriented flow (TOFlow)\protect\footnote{Code: https://github.com/anchen1011/toflow} \citep{TOFlow} is shown in Fig.\ \ref{figs1009}. TOFlow combines the network for optical flow estimation with the reconstruction network, and trains them jointly to obtain optical flow network tailored to a specific task such as video SR, video interpolation and video deblurring.

TOFlow adopts SpyNet \citep{Ranjan_2017_CVPR} as the network for the optical flow estimation, and then adopts a spatial transformer network (STN) to warp the neighboring frame according to the computed optical flow. Then the final result is obtained by an image processing network. For the video super-resolution task, the image processing module consists of 4 convolutional layers, where kernel sizes are 9$\times$9, 9$\times$9, 1$\times$1, and 1$\times$1, respectively, and the numbers of channels are 64, 64, 64, and 3, respectively.

\begin{figure}[!htbp]
\centering
\includegraphics[width=0.79\columnwidth]{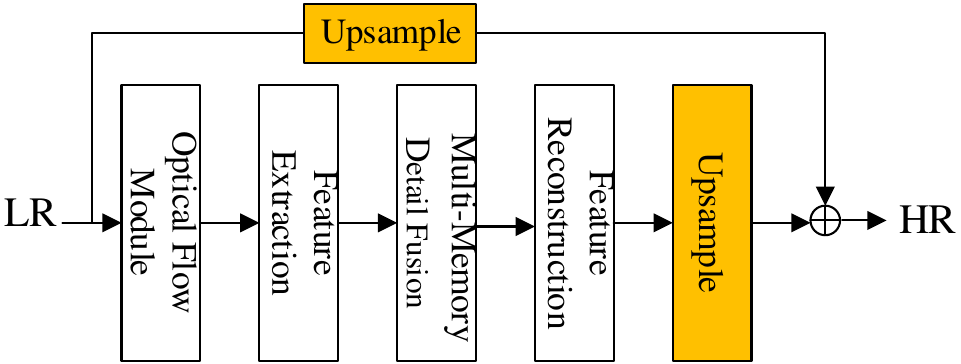}
\caption{The network architecture of MMCNN \citep{MMCNN}.}
\label{figs1010}
\end{figure}

\subsubsection{MMCNN} 
The architecture of the multi-memory convolutional neural network (MMCNN)\protect\footnote{Code: https://github.com/psychopa4/MMCNN} \citep{MMCNN} is shown in Fig.\ \ref{figs1010}, and it consists of 5 major modules: optical flow module for motion estimation and motion compensation, feature extraction, multi-memory detail fusion, feature reconstruction, and upsample modules, where the last module uses a sub-pixel convolutional layer.

Consecutive input frames are first processed by the optical flow estimation module to make neighboring frames align with the target frame and then the warped frames are fed into subsequent network modules to attain the residual image of the target frame. Finally, this residual image is added into the upsampled LR target frame, which is computed by bicubic interpolation, to obtain the super-resolution result. In the multi-memory detail fusion module, MMCNN adopts the ConvLSTM module \citep{convLSTM} to merge the spatio-temporal information. Moreover, the feature extraction, detail fusion, and feature reconstruction modules are all built based on residual dense blocks \citep{RDN, Dense}, where the key difference among them is merely the type of network layers.

\subsubsection{MEMC-Net}  
\begin{figure}[!htbp]
\centering
\includegraphics[width=0.95\columnwidth]{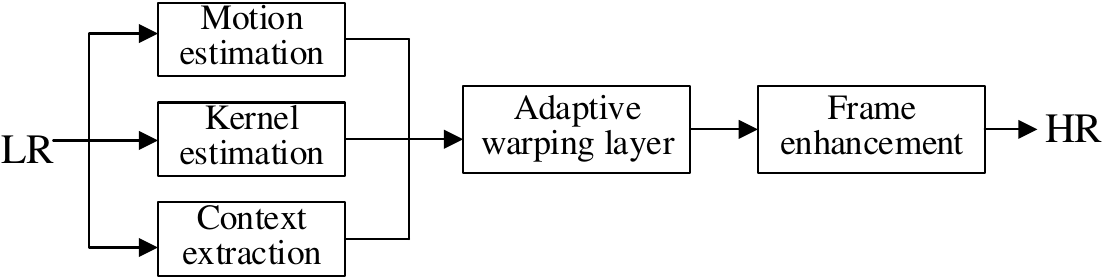}
\caption{The network architecture of MEMC-Net \citep{MEMC-Net}.}
\label{figs1012}
\end{figure}

The motion estimation and motion compensation network (MEMC-Net)\protect\footnote{Code: https://github.com/baowenbo/MEMC-Net} \citep{MEMC-Net}, as shown in Fig.\ \ref{figs1012}, mainly proposes an adaptive warping layer.

The adaptive warping layer warps the neighboring frame through the estimated optical flow and the convolutional kernel, which are resulted from a  motion estimation network and a kernel estimation network, respectively, and aligns the neighboring frame with the target frame. The motion estimation network adopts FlowNet \citep{FlowNet}, and the kernel estimation network uses an improved U-Net \citep{u-net} including five max-pooling layers, five un-pooling layers and skip connections from the encoder to the decoder. In MEMC-Net, the architecture of the super-resolution module, namely frame enhancement module, is similar to that of EDSR \citep{EDSR}. In order to deal with the occlusion problem, it adopts a pre-trained ResNet18 \citep{ResNet} to extract the feature of input frames. Moreover, it feeds the output of the first convolutional layer of ResNet18 as the context information into the adaptive warping layer to perform the same operation.

\begin{figure}[!ht]
\centering
\includegraphics[width=0.569\columnwidth]{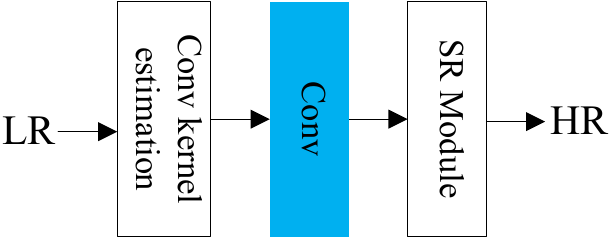}
\caption{The network architecture of RTVSR \citep{RTVSR}. Here SR Module denotes super-resolution module.}
\label{figs1013}
\end{figure}

\subsubsection{RTVSR}  
The real-time video super-resolution (RTVSR) \citep{RTVSR}, as shown in Fig.\ \ref{figs1013}, adopts a convolutional network called motion convolutional kernel estimation network, which is a full convolution codec structure, to estimate the motion between the target frame and the neighboring frame and produce a pair of 1D convolutional kernel corresponding to the current target frame and neighboring frame. Then the neighboring frame is warped by using estimated convolutional kernels to make it align with the target frame. RTVSR designs an important component called gated enhance units (GEUs) to learn useful features, which is an improved variant based on \citep{8462170}.

\begin{figure}[!htbp]
\centering
\includegraphics[width=0.998\columnwidth]{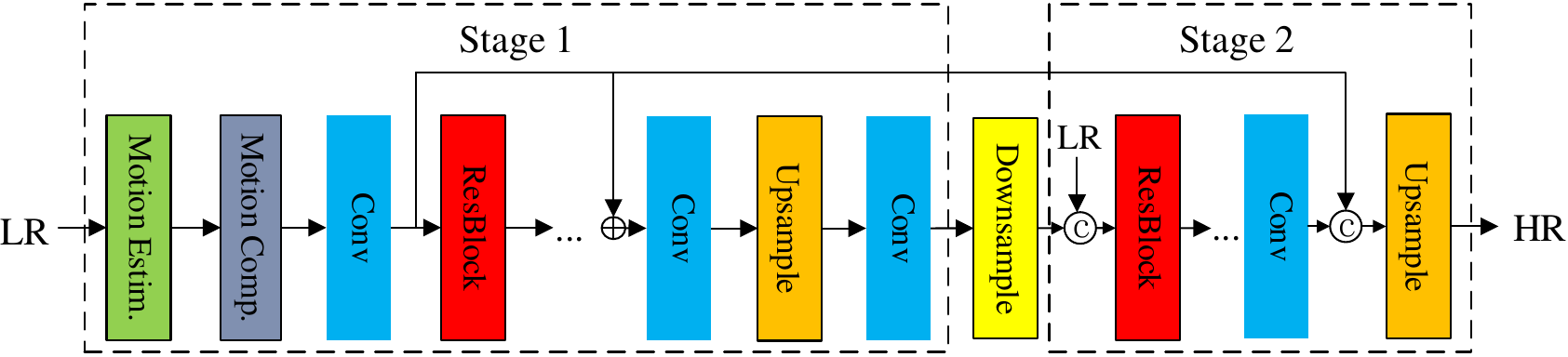}
\caption{The network architecture of MultiBoot VSR \citep{MultiBoot}.}
\label{figs1014}
\end{figure}
	
\subsubsection{MultiBoot VSR}  
The multi-stage multi-reference bootstrapping for video super-resolution (MultiBoot VSR) \citep{MultiBoot} consists of two stages. That is, in order to further improve performance, the output of the first stage is used as the input of the second stage. The network architecture of MultiBoot VSR is shown in Fig.\ \ref{figs1014}.

The LR frames are input to the FlowNet 2.0 to compute optical flow and perform the motion compensation. Then the processed frames are fed into the first-stage network to attain the super-resolution result of the target frame. In the second stage of MultiBoot VSR, the output from the previous stage is downsampled, concatenated with the initial LR frame, and then input to the network to obtain final super-resolution result for the target frame.

\begin{figure}[!htbp]
\centering
\includegraphics[width=0.657\columnwidth]{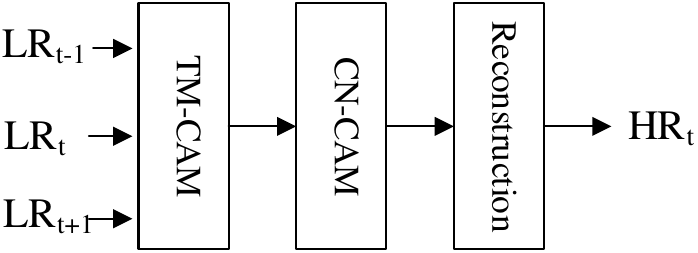}
\caption{The network architecture of MuCAN \citep{MuCAN}.}
\label{figmucan}
\end{figure}
\subsubsection{MuCAN}
The architecture of multi-correspondence aggregation network (MuCAN) \citep{MuCAN} is shown in Fig. \ref{figmucan}. MuCAN is an end-to-end network consisting of a temporal multi-correspondence aggregation module (TM-CAM), a cross-scale non-local-correspondence aggregation module (CN-CAM), and a reconstruction module.

In TM-CAM, two neighboring LR frames are first encoded into lower-resolution features to be more stable and robust to noise. Then the aggregation starts in the original LR feature space by an aggregation unit (AU) to compensate large motion while progressively moving up to low-level/high-resolution stages for subtle sub-pixel shift. In a single AU, a patch-based matching strategy is used since it naturally contains structural information. Multiple candidates are then aggregated to obtain sufficient context information. The aggregated information is then passed to CN-CAM, which then uses a pyramid structure based on AvgPool to execute spatio-temporal non-local attention and coarse-to-fine spatial attention. Finally, the results are aggregated and sent to the reconstruction module to yield the final HR result.

\begin{figure}[!htbp]
\centering
\includegraphics[width=0.795\columnwidth]{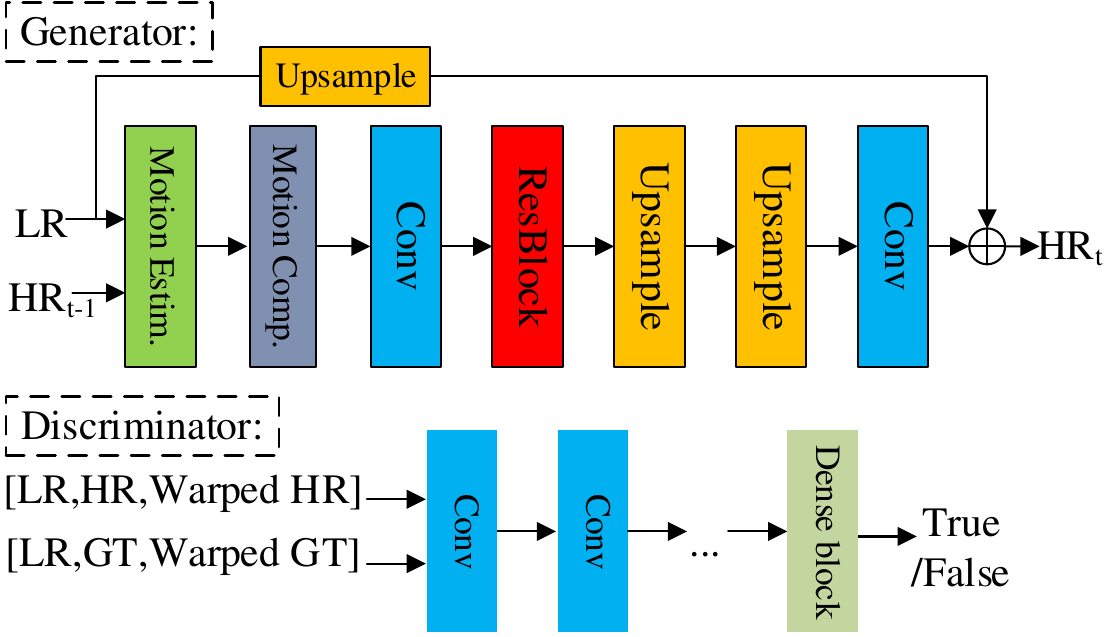}
\caption{The network architecture of TecoGAN \citep{TecoGAN}.}
\label{figs1008}
\end{figure}

\subsubsection{TecoGAN}  
Temporally coherent GAN (TecoGAN)\protect\footnote{Code: https://github.com/thunil/TecoGAN} \citep{TecoGAN} mainly proposes a spatio-temporal discriminator for realistic and coherent video super-resolution, and a novel ``Ping-Pong" loss to tackle recurrent artifacts. Like GAN, TecoGAN also consists of a generator and a discriminator and its architecture is shown in Fig.\ \ref{figs1008}.

The generator takes the target frame, the previous frame and previous estimated HR frames as inputs. First, input frames are fed into the optical flow module, which is a CNN similar to the optical flow estimation module in FRVSR \citep{FRVSR}. In this module, the LR optical flow between the target frame and neighboring frames is estimated and enlarged by the bicubic interpolation to attain the corresponding HR optical flow. Then the previous HR frame is warped by the HR optical flow. The warped previous HR frame and target frame are fed into subsequent convolutional modules that include two convolutional layers, a residual block and two upsample modules with a deconvolution layer, to yield a restored target frame. Moreover, the discriminator assesses the quality of super-resolution results. The discriminator takes the generated results and GT as inputs, where each of them has three components, that is, three consecutive HR frames, three corresponding upsampled LR frames and three warped HR frames. With such input formats, the spatial over-smooth and temporal inconsistence in the final results can be relieved. TecoGAN also proposes a ``ping-pong" loss function to reduce the long-term temporal detail drift and make super-resolution results more natural.


\begin{figure}[!htbp]
\centering
\includegraphics[width=0.75\columnwidth]{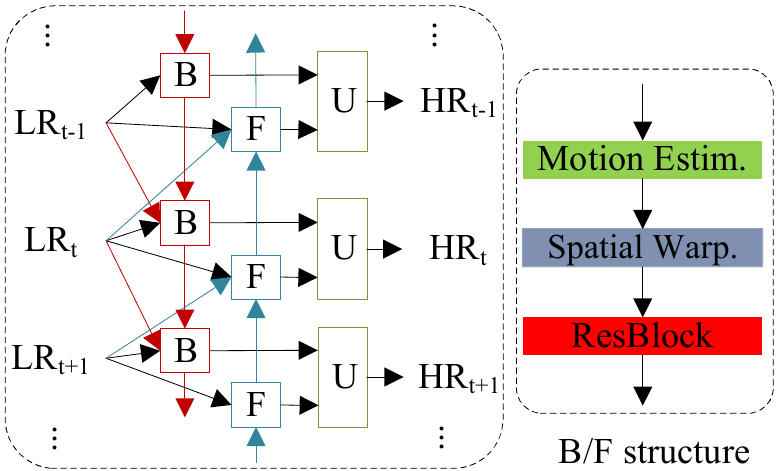}
\caption{The network architecture of BasicVSR \citep{chan2020basicvsr}.}
\label{figBasicVSR}
\end{figure}

\subsubsection{BasicVSR}  
The authors proposed a generic framework for video super resolution, called BasicVSR, as shown in Fig. \ref{figBasicVSR}. It is a typical
bidirectional recurrent network, which mainly consists of three modules: the backward (B) module, the forward (F) module, and the upsampling (U) module. The B module receives the output of the next B module, current frame, and the following frame, while the F module receives the output of the previous F module, current frame, and the preceding frame. Then the outputs of the two modules are fused through a U module to yield the super-resolved current frame. These processes iterate until all the frames are super-resolved. The B/F module composes of generic components: the motion estimation, spatial warping, and residual blocks. The authors further propose two processing mechanisms the information-refill and coupled propagation, which consist of the IconVSR algorithm. The former addresses the performance degradation caused by misalignment, and the latter deals with the lack of information interaction between the forward processing and the backward processing in BasicVSR. In the information-refill mechanism, if the currently processed frame is in the selected keyframe set, it will be fused; otherwise, the aligned result will be directly sent into the residual block without fusion. This mechanism relieves error accumulation caused by misalignment, thus avoiding the performance degradation. In the coupling propagation mechanism, the output of backward propagation is directly used as the input of forward propagation, so as to achieve information interaction between them.

\textbf{In summary}, the MEMC techniques are used to align neighboring frames with a target frame, and are probably the most common method for solving video super-resolution tasks. However, the problem is that they cannot guarantee the accuracy of motion estimation when lighting changes dramatically or there are large motions in videos. In these cases, the performance of the video super-resolution degrades greatly. This is confirmed by the assumption in \citep{Iterative1981}. When dealing with complex motions (not only large motions) and varying illumination, the calculation of motion estimation based on optical flow methods may break the hypothesis of brightness consistency, small moti on, and spatial coherence. Then the estimation of optical flow becomes inaccurate, and there arises errors, which easily results in artifacts and blurring. To address this issue, the methods with alignment (e.g., the deformable convolution which is presented as one module in the deep network to align frames) and the methods without alignment are both proposed.
\begin{figure}[!htbp]
\centering
\includegraphics[scale=0.616]{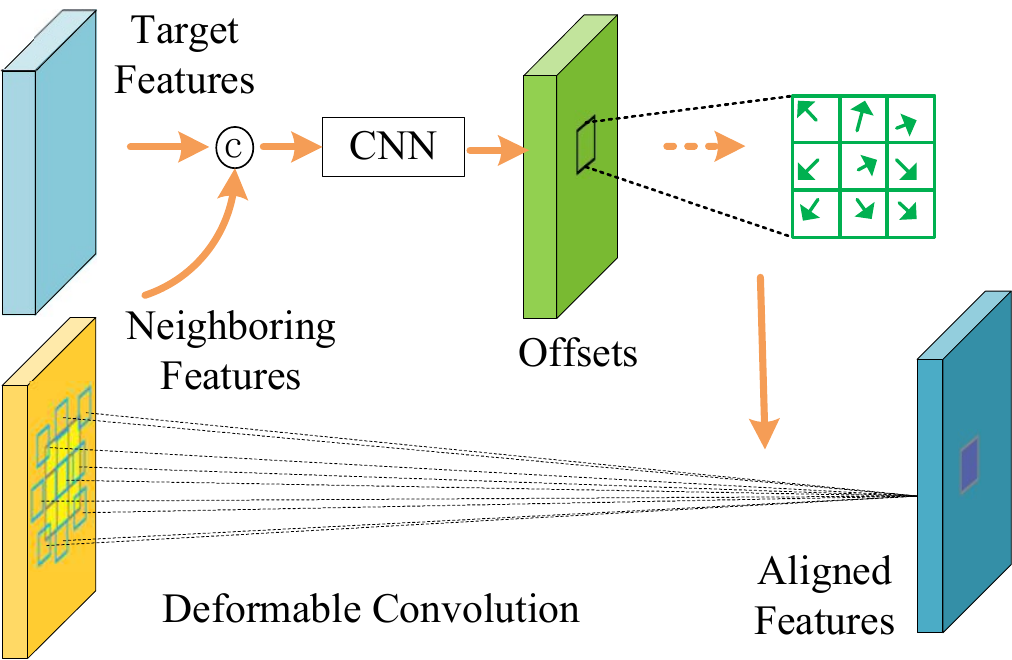}
\caption{Deformable convolution for frame alignment.}
\label{dcn}
\end{figure}

\subsection{Deformable Convolution Methods}
The deformable convolutional network was first proposed by \citet{DCN} and the improved variant \citep{DCNV2} was proposed in 2019. In ordinary CNNs, the convention is to use a fixed geometric structure in a layer, which restricts the network's capability to model geometric transformations. In contrast, the deformable convolution is able to overcome this limitation. The illustration of the deformable convolution for feature alignment is shown in Fig.\ \ref{dcn}. The target feature maps concatenating with the neighboring  feature maps are projected to attain offsets via additional convolutional layers. The offsets are applied to the conventional convolution kernel to yield a deformable convolution kernel, and then it is convolved with the input feature maps to produce the output feature maps. The methods that adopt deformable convolution mainly include the enhanced deformable video restoration (EDVR) \citep{EDVR}, deformable non-local network (DNLN) \citep{DNLN}, and temporally deformable alignment network (TDAN) \citep{TDAN}, which are depicted in detail as follows.

\begin{figure}[!htbp]
\centering
\includegraphics[width=0.67\columnwidth]{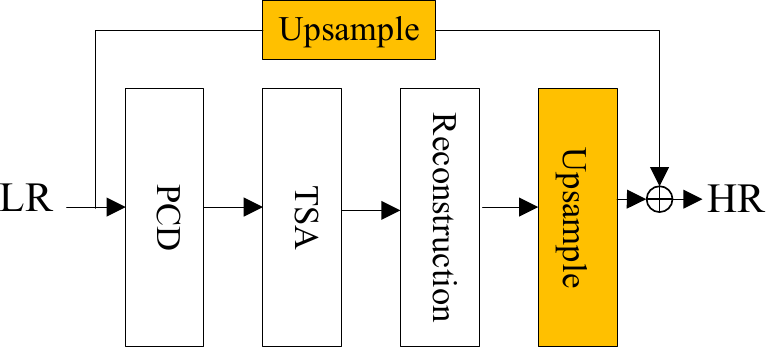}
\caption{The network architecture of EDVR \citep{EDVR}, where PCD is the pyramid, cascading and deformable alignment module, and TSA is the temporal-spatial attention fusion module.}
\label{figs1015}
\end{figure}

\subsubsection{EDVR }
The enhanced deformable video restoration (EDVR)\protect\footnote{Code: https://github.com/xinntao/EDVR} \citep{EDVR}, as shown in Fig.\ \ref{figs1015}, is the champion model in the NTIRE19 Challenge \citep{Nah_2019_CVPR_Workshops1,Nah_2019_CVPR_Workshops2}. EDVR proposes two key modules: the pyramid, cascading and deformable (PCD) alignment module as in \citep{Ranjan_2017_CVPR,PWC-Net,LiteFlowNet,LightweightOF} and the temporal-spatial attention (TSA) fusion module, which are used to solve large motions in videos and to effectively fuse multiple frames, respectively.

EDVR mainly consists of four parts: one PCD alignment module, a TSA fusion module, a reconstruction module, and an upsample module using a sub-pixel convolutional layer. Firstly, the input frames are aligned by the PCD alignment module, and then the aligned frames are fused by the TSA fusion module. Then the fused results are fed into the reconstruction module to refine the features, and then through the up-sampling, a HR image called the residual image is obtained. The final result is obtained by adding the residual image to a direct upsampling target frame. To further improve performance, EDVR also adopts a two-phase approach, whose second phase is similar to the first but with a shallower network depth.


\begin{figure}[!htbp]
\centering
\includegraphics[width=0.77\columnwidth]{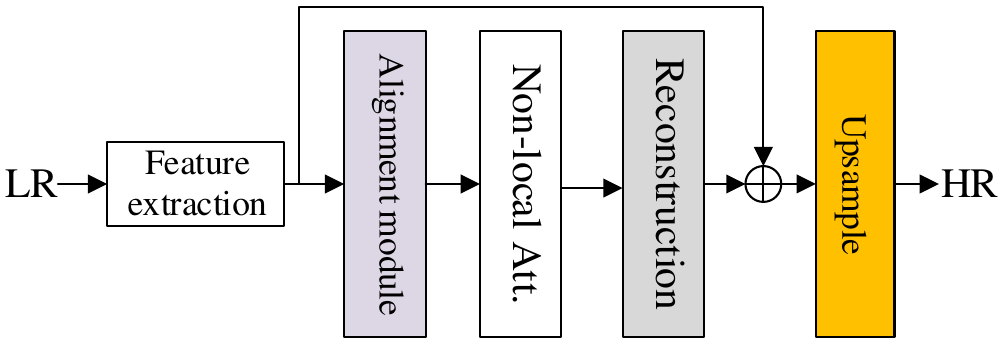}
\caption{The network architecture of DNLN \citep{DNLN}. Here Non-local Att. is the non-local attention module.}
\label{figs1016}
\end{figure}

\subsubsection{DNLN}
The deformable non-local network (DNLN)\protect\footnote{Code: https://github.com/wh1h/DNLN} \citep{DNLN}, as shown in Fig.\ \ref{figs1016}, designs an alignment module and a non-local attention module based on the deformable convolution \citep{DCN,DCNV2} and non-local networks \citep{NLN}, respectively. The alignment module uses the hierarchical feature fusion module (HFFB) \citep{2019arXiv190710399H} within the original deformable convolution to generate convolutional parameters. Moreover, DNLN utilizes multiple deformable convolutions in a cascaded way, which makes inter-frame alignment more accurate.



\begin{figure}[!htbp]
\centering
\includegraphics[width=0.66\columnwidth]{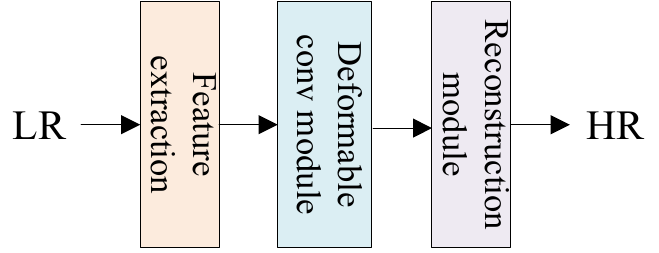}
\caption{The network architecture of TDAN \citep{TDAN}.}
\label{figs1017}
\end{figure}

\subsubsection{TDAN}
The temporally deformable alignment network (TDAN)\protect\footnote{Code: https://github.com/YapengTian/TDAN-VSR-CVPR-2020} \citep{TDAN}, as shown in Fig.\ \ref{figs1017}, applies deformable convolution to the target frame and the neighboring frame, and attains corresponding offsets. Then the neighboring frame is warped in terms of the offsets to align with the target frame. TDAN is divided into three parts, i.e., a feature extraction module, a deformable convolution module and a reconstruction module.

\begin{figure}[!htbp]
\centering
\includegraphics[width=0.926\columnwidth]{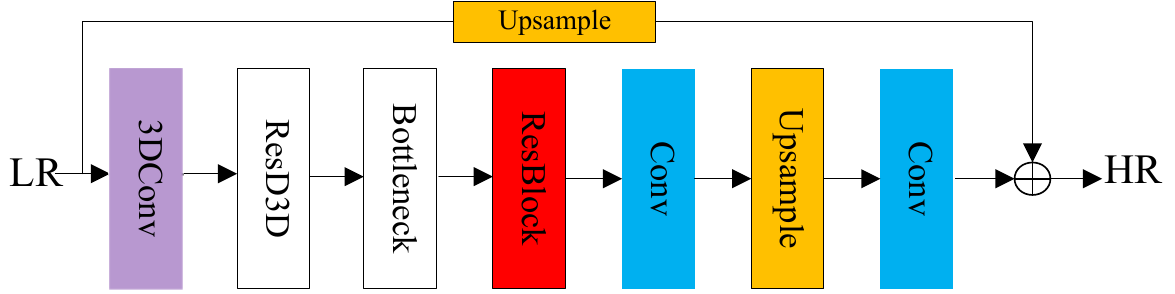}
\caption{The network architecture of D3Dnet \citep{D3Dnet}.}
\label{figsd3d}
\end{figure}

\subsubsection{D3Dnet}
The architecture of the deformable 3D convolution network (D3Dnet)\protect\footnote{Code: https://github.com/XinyiYing/D3Dnet} \citep{D3Dnet} is shown in Fig. \ref{figsd3d}. D3Dnet proposes 3D deformable convolution to achieve strong spatio-temporal feature modeling capability. The inputs are first fed to a 3D convolutional layer to generate features, which are then fed to 5 Residual Deformable 3D Convolution (ResD3D) blocks to achieve motion compensation and capture spatial information.


\begin{figure}[!htbp]
\centering
\includegraphics[width=0.999\columnwidth]{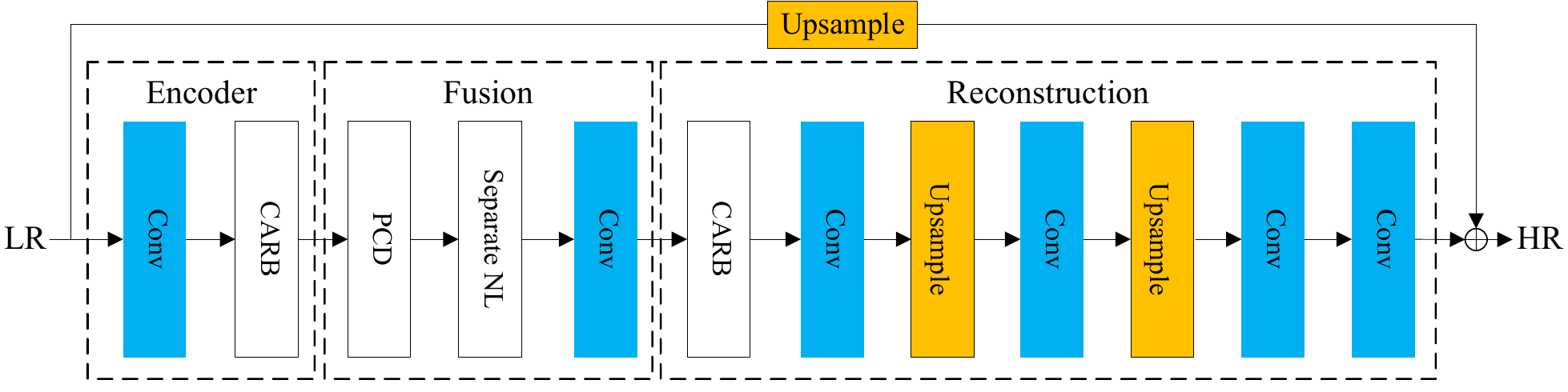}
\caption{The architecture of VESR-Net \citep{chen2020vesr}, where CARB is the channel-attention residual block, and Separate NL denotes the separate non-local architecture.}
\label{figvesr}
\end{figure}

\subsubsection{VESR-Net}
The architecture of video enhancement and super-resolution network (VESR-Net) \citep{chen2020vesr}, as shown in Fig. \ref{figvesr}, is the champion model in the Youku video enhancement and super-resolution challenge. VESR-Net mainly consists of a feature encoder, a fusion module and a reconstruction module.

The LR frames are firstly processed by the feature encoder consisting of a convolution layer and several stacked channel-attention residual blocks (CARBs) \citep{RCAN}. Then in the fusion module, the PCD convolution in \citep{EDVR} performs the inter-frame feature alignment. The separate non-local submodule (Separate NL) divides feature maps in spatial, channel and temporal dimensions and processes them to obtain correlation information separately. In contrast to the vanilla non-local \citep{NLN} architecture, Separate NL can fuse the information across video frames and across pixels in each frame with less parameters and shallower network. Finally, VESR-Net utilizes CARBs followed with a feature decoder for upsampling in the reconstruction module, where the upsample module is implemented by a sub-pixel convolutional layer. And it outputs the super-resolved frame by adding with the bicubic-interpolation LR target frame.




\textbf{The evolution of methods with alignment.} In the methods with alignment, the motion estimation and motion compensation techniques, as a classic research topic in computer vision, have been applied to video super-resolution in the early years. MEMC has wide range of applications such as video coding and enhancing the interlaced scanning. As the advent of deep learning based VSR, many works employ MEMC to capture the motion information contained in video frames. The early work of MEMC is Deep-DE \citep{Deep-DE}, and some recently proposed methods such as VESPCN \citep{VESPCN}, SOFVSR \citep{SOFVSR}, TOFlow \citep{TOFlow} and FRVSR \citep{FRVSR} also adopted MEMC techniques. Specifically, early video super-resolution algorithms adopt traditional MEMC methods such as Druleas in VSRnet \citep{VSRnet}, while subsequent algorithms such as VESPCN \citep{VESPCN}, TOFlow \citep{TOFlow} and FRVSR \citep{FRVSR} mainly design sub-module or sub-network for MEMC.

However, the accuracy of most MEMC methods is usually not guaranteed. When the luminance changes or the videos contain large motions between frames, the performance of VSR degrades dramatically. Hence, the deformable convolution (DConv), which is not sensitive to varying lighting and motion conditions, has attracted more attention from researchers. DConv applies a learnable offset to each sampling point compared with the conventional convolution. Therefore, DConv can not only expand the receptive field of convolution kernel, but also enrich the shape of receptive field. When handling varying lighting and motion conditions, the conventional convolution with fixed kernel and limited receptive field may not be capable of capturing varying conditions. While DConv uses a learnable parameter for the kernel to analyze lighting and motion features, which can better capture complex motions and illumination changes. The deformable convolution was proposed by \citet{DCN} to enhance the transformation modeling capability of CNNs for the geometric variations of objects. In the VSR methods, TDAN \citep{TDAN} first utilized it to perform inter-frame alignment. After that, DNLN \citep{DNLN}, EDVR \citep{EDVR}, and D3Dnet \citep{D3Dnet} further promote it for frame alignment. Nevertheless, the deformable convolution still has some drawbacks including high computational complexity and harsh convergence conditions. Therefore, there is still room for improvement of this technique in the future.

In addition, the performance of the MEMC-based methods will degrade greatly when there were dramatic lighting changes and large motions in videos. Nevertheless, the network architecture is one of the important factors to affect its performance. Other factors include the training dataset, training strategy, data preprocessing, hyper-parameter setting, iteration times, etc. Although the MEMC-based methods have the limitation to deal with videos containing lighting changes and large motions, they can be counteracted by other network designs and training settings. For example, BasicVSR/IconVSR adopts a bidirectional recurrent network as backbone, which fully utilizes the global information from the video sequences and expands receptive field. Thus, they may gain superior performance compared with the other MEMC methods, which mainly use convolutions. Moreover, the training process, which uses a Cosine annealing scheme ~\citep{loshchilov10sgdr}, is probably more refined.

\section{Methods without Alignment}
In contrast to the methods with alignment, the methods without alignment do not align neighboring frames for video super-resolution. This type of methods mainly exploit the spatial or spatio-temporal information for feature extraction. According to the dominating techniques utilized for initial feature extraction, we further categorize them into five types: the 2D convolution methods (2D Conv), 3D convolution methods (3D Conv), recurrent convolutional neural network (RCNN), non-local network based, and other methods. Among them, the first type falls into the spatial methods, while the following three are the spatio-temporal methods, whose characteristic is to exploit both the spatial and temporal information from input videos. Other methods include the ones do not belong to any of the former. We present them in detail as follows.

\subsection{2D Convolution Methods}
Instead of alignment operations such as motion estimation and motion compensation between frames, the input frames are directly fed into a 2D convolutional network to spatially perform feature extraction, fusion and super-resolution operations. This may be a simple approach for solving the video super-resolution problem since it makes the network learn the correlation information within frames by itself. The representative methods are VSRResFeatGAN \citep{VSRResNet} and FFCVSR \citep{FFCVSR}.

\begin{figure}[!htbp]
\centering
\includegraphics[width=0.786\columnwidth]{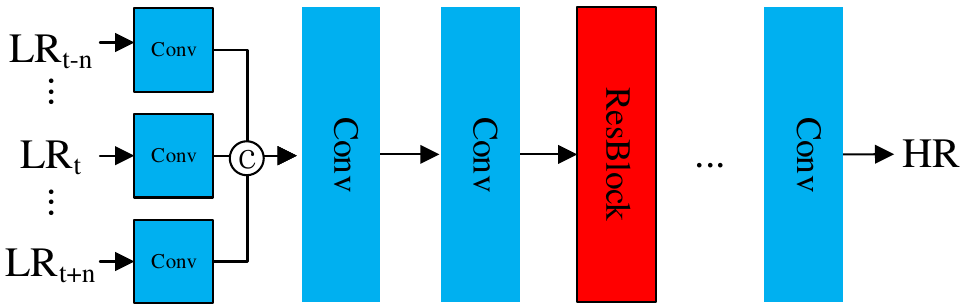}
\caption{The architecture of the generator in VSRResFeatGAN \citep{VSRResNet}.}
\label{figs1018}
\end{figure}
\subsubsection{VSRResFeatGAN}
VSRResFeatGAN \citep{VSRResNet} utilizes GAN to address VSR tasks  and find a good solution by adversarial training. The generator shown in Fig.\ \ref{figs1018} consists of convolutional layers and residual blocks. And each residual block is composed of two convolutional layers and is followed by a ReLU activation function. Moreover, the discriminator consists of three groups of convolutions and a fully connected layer, where each group includes a convolutional layer, Batch Normalization (BN), and LeakyReLU. The discriminator determines whether the output of the generator is a generated image or GT image. Then the result of the discriminator reacts to the generator, and promotes it to yield results closer to the GT images. Finally, a relative satisfactory solution is obtained through an iterative optimization.




\begin{figure}[!htbp]
\centering
\includegraphics[width=0.493\columnwidth]{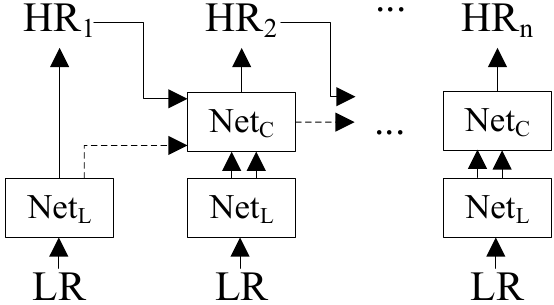}
\caption{The architecture of FFCVSR \citep{FFCVSR}. Here Net$_{C}$ is the context network, and Net$_{L}$ is the local network.}
\label{figs1019}
\end{figure}

\subsubsection{FFCVSR}
The architecture of the frame and feature-context video super-resolution (FFCVSR)\protect\footnote{Code: https://github.com/linchuming/FFCVSR} \citep{FFCVSR} is shown in Fig.\ \ref{figs1019}. Unlike common MEMC techniques, FFCVSR consists of several local networks and context networks and utilizes inter-frame information in a different way. The LR unaligned video frames and the HR output of the previous frame are directly taken as inputs to the network for the purpose of restoring high-frequency details and maintaining temporal consistency.



\textbf{In summary}, the above two methods both exploit spatial correlation between frames for VSR tasks. VSRResFeatGAN utilizes adversarial training of GANs to find an appropriate solution. As the discriminator in GANs has to guess whether the generated frame is close to the ground truth, the VSR results in terms of PSNR and SSIM are not always satisfactory compared with other methods, such as FFCVSR.

\subsection{3D Convolution Methods}
The 3D convolutional module \citep{3D,3D1} operates on spatio-temporal domain, compared with 2D convolution, which only utilizes spatial information through the sliding kernel over input frame. This is beneficial to the processing of video sequences, as the correlations among frames are considered by extracting temporal information. The representative 3D convolution methods for VSR are DUF \citep{DUF}, FSTRN \citep{FSTRN}, and 3DSRnet \citep{3DSRnet}.
\begin{figure}[!htbp]
\centering
\includegraphics[width=0.97\columnwidth]{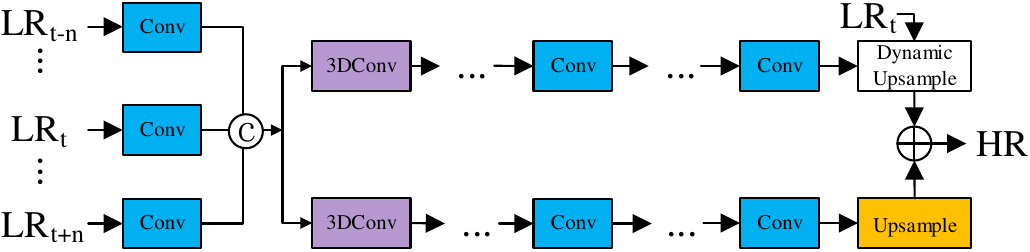}
\caption{The network architecture of DUF \citep{DUF}.}
\label{figs1020}
\end{figure}

\subsubsection{DUF}
 The dynamic upsampling filters (DUF)\protect\footnote{Code: https://github.com/yhjo09/VSR-DUF} \citep{DUF} has been proposed, as shown in Fig.\ \ref{figs1020}. It is inspired by the dynamic filter network \citep{DFN} that can generate corresponding filters for specific inputs and then apply them to generate corresponding feature maps.

The structure of the dynamic up-sampling filter, together with the spatio-temporal information learned by 3D convolution, can avoid the use of motion estimation and motion compensation. DUF performs not only filtering, but also the up-sampling operation. In order to enhance high-frequency details of the super-resolution result, DUF uses a network to estimate residual map for the target frame. The final result is the sum of the residual map and the LR target frame processed by the dynamic upsample module with learned filters.

\begin{figure}[!htbp]
\centering
\includegraphics[width=0.95\columnwidth]{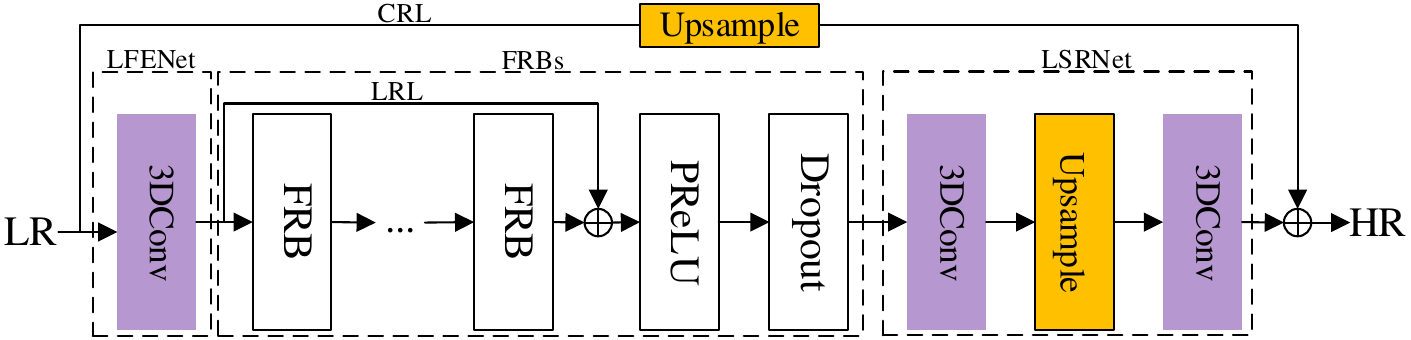}
\caption{The network architecture of FSTRN \citep{FSTRN}. Here FRB denotes the fast spatio-temporal residual block.}
\label{figs1021}
\end{figure}

\subsubsection{FSTRN}
The fast spatio-temporal residual network (FSTRN) \citep{FSTRN} uses a factorized 3D convolution to extract information contained in consecutive frames, as shown in Fig.\ \ref{figs1021}. In FSTRN, a $k\times k\times k$ 3D convolutional kernel is decomposed into 2 cascaded kernels, whose sizes are $1\times k \times k$ and $k\times1\times1$, respectively, to reduce the computation caused by directly using the 3D convolution.

FSTRN consists of the following four parts: an LR video shallow feature extraction net (LFENet), fast spatio-temporal residual blocks (FRBs), an LR feature fusion and up-sampling SR net (LSRNet), and a global residual learning (GRL) module. The GRL is mainly composed of two parts: LR space residual learning (LRL) and cross-space residual learning (CRL). The LRL is introduced along with the FRBs. And the CRL directly maps the LR video to the HR space. The designs of CRL and LRL can communicate the LR and HR space. Besides, FSTRN adopts a dropout layer after LRL to enhance generalization ability of the network. LFENet using 3D convolution to extract features for consecutive LR input frames. FRBs, including the decomposed 3D convolutional layers, are responsible for extracting spatio-temporal information contained in input frames. LSRNet is used to fuse information from previous layers and conducting up-sampling.

\begin{figure}[!htbp]
\centering
\includegraphics[width=0.86\columnwidth]{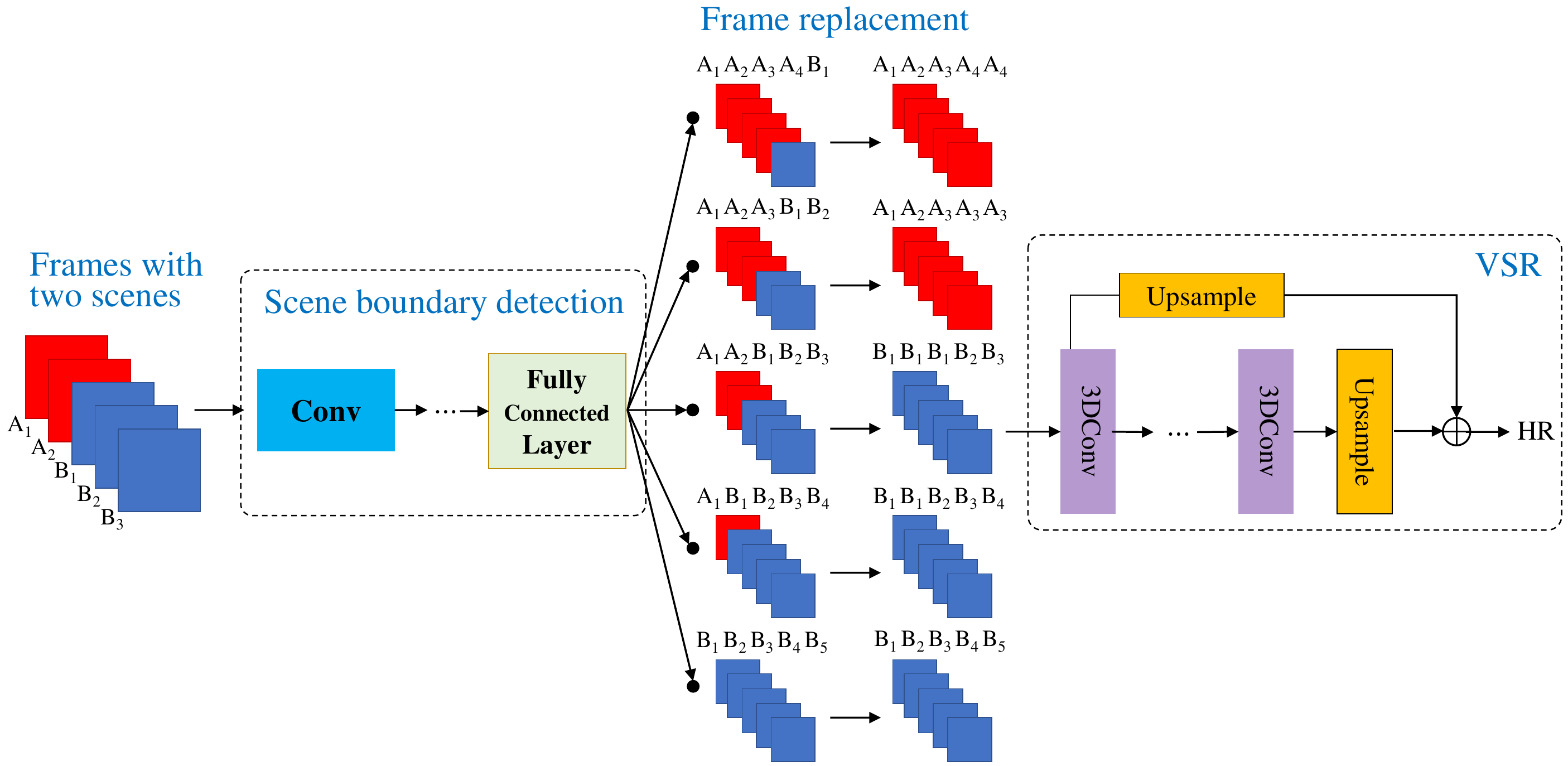}
\caption{The network architecture of 3DSRNet \citep{3DSRnet}.}
\label{figs1022}
\end{figure}

\subsubsection{3DSRNet}
The 3D super-resolution network (3DSRNet)\protect\footnote{Code: https://github.com/sooyekim/3DSRnet} \citep{3DSRnet} uses 3D convolution to extract spatio-temporal information contained in consecutive frames for VSR tasks. The network architecture is shown in Fig.\ \ref{figs1022}. The sub net of 3DSRNet can preprocess scene change as shown in the figure. When frames of five different scenes getting involved into convolution, the sub net classifies the exact location of the scene boundary through the module of scene boundary detection, and replaces the different scene frames with the temporally closest frame of the same scene as the current middle frame. Finally, the updated five frames are sent for subsequent video super-resolution sub network. This approach overcomes performance degradation caused by scene change to some extent.
\begin{figure}[!htbp]
\centering
\includegraphics[width=0.75\columnwidth]{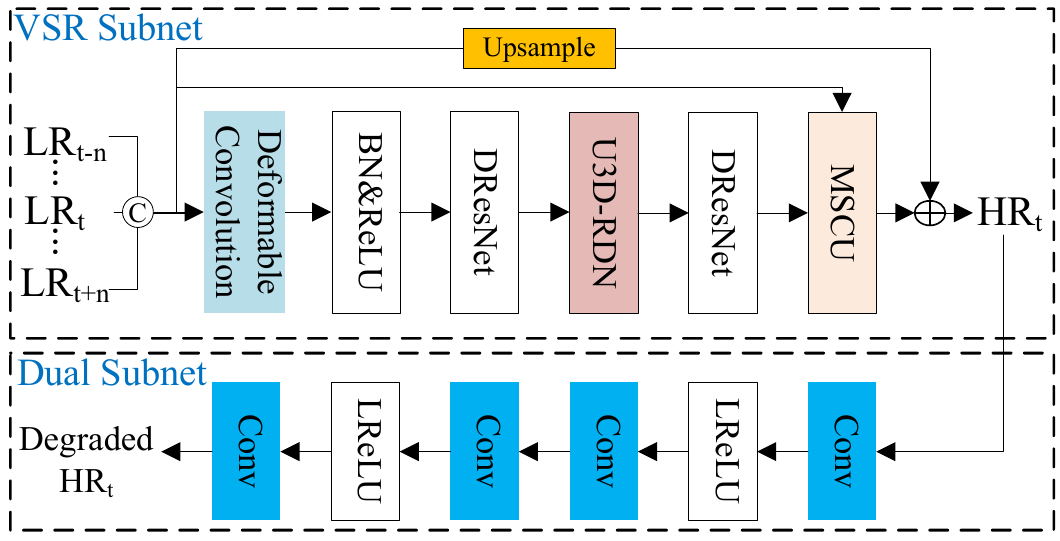}
\caption{The network architecture of DSMC \citep{DSMC2021}.}
\label{figDSMC}
\end{figure}
\subsubsection{DSMC}
A deep neural network with Dual Subnet and Multi-stage Communicated Upsampling (DSMC)\protect\footnote{Code: https://github.com/iPrayerr/DSMC-VSR} \citep{DSMC2021} is proposed for super-resolution of videos with large motion. The architecture is shown in Fig. \ref{figDSMC}. It designs a U-shaped residual dense network with 3D convolution (U3D-RDN) for fine implicit MEMC as well as coarse spatial feature extraction. moreover, DSMC presents a new Multi-Stage Communicated Upsampling (MSCU) module to make full use of the intermediate results of upsampling for guiding the VSR. Besides, a dual subnet is devised to aid the training of DSMC, whose dual loss helps to reduce the solution space and enhance the generalization ability.

DSMC firstly performs deformable convolution on input consecutive frames for coarse feature extraction. The output feature maps are then processed by a deformable residual network (DResNet) \citep{lei2018temporal} to extract fine spatial information. Next, the feature maps are input to U3D-RDN for dimension reduction and correlation analyzation of spatio-temporal feature. Followed by another DResNet module, the feature maps are sent to MSCU module. Finally, with the aid of a dual subnet for training, DSMC yields the super-resolved HR frames. It is noted that only the output of the dual subnet and the result of VSR subnet are used for the loss computation of DSMC.

\textbf{In brief}, these 3D convolutional methods can extract spatio-temporal correlations contained in consecutive frames, rather than perform the motion estimation to extract motion information contained in frames and motion compensation to align them. However, most of the methods have relatively higher computational complexities compared with those of 2D convolutional methods, which limits them for real-time video super-resolution tasks.

\subsection{Recurrent Convolutional Neural Networks (RCNNs)}
It is well known that RCNNs have strong capacity in modelling temporal dependency in sequential data, e.g., natural language, video and audio. A straightforward way is to use RCNNs to handle video sequences. Based on this key idea, several RCNN methods such as BRCN \citep{NIPS2015_5778, BRCN}, STCN \citep{STCN2017AAAI},  and RISTN \citep{RISTN} have been proposed for video super-resolution.

\begin{figure}[!htbp]
\centering
\includegraphics[width=0.85\columnwidth]{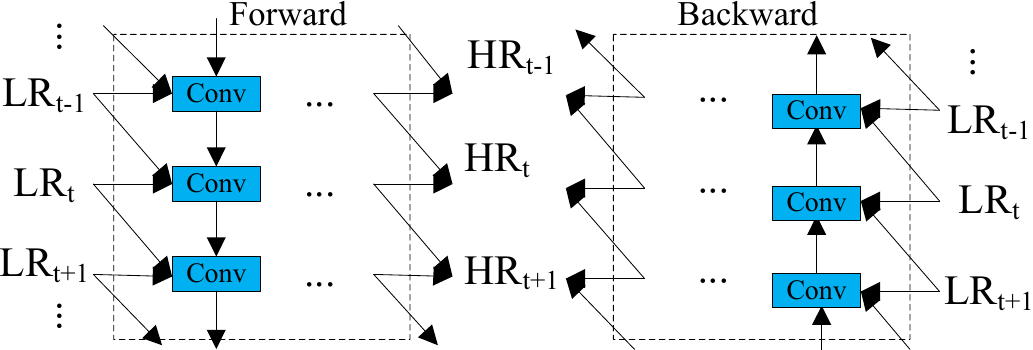}
\caption{The network architecture of BRCN \citep{NIPS2015_5778, BRCN}.}
\label{figs1024}
\end{figure}

\subsubsection{BRCN}
The bidirectional recurrent convolutional network (BRCN) \citep{NIPS2015_5778, BRCN}, as shown in Fig.\ \ref{figs1024}, is composed of two modules: a forward sub-network and a backward one with a similar structure, which only differ in the order of processing sequence. The forward subnet is responsible for modeling the temporal dependency from previous frames, while the backward subnet models temporal dependency from subsequent frames.



\begin{figure}[!htbp]
\centering
\includegraphics[width=0.77\columnwidth]{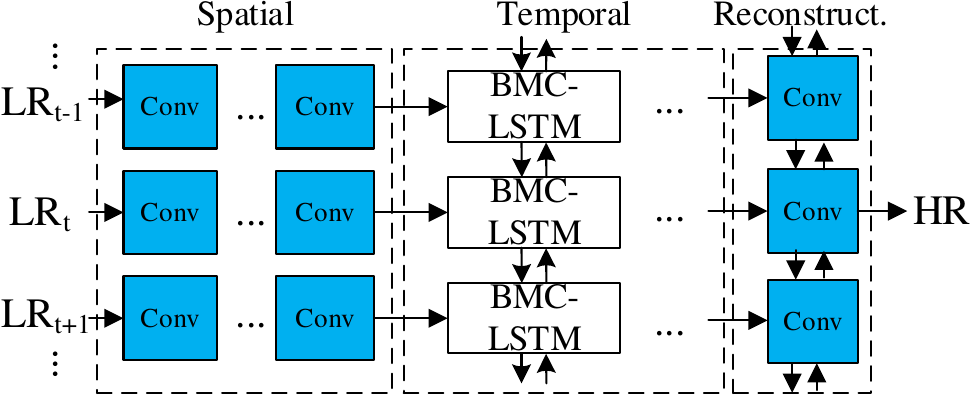}
\caption{The network architecture of STCN \citep{STCN2107AAAI}. Here BMC denotes the bidirectional multi-scale convolution.}
\label{figs1023}
\end{figure}	

\subsubsection{STCN}
The spatio-temporal convolutional network (STCN) \citep{STCN2017AAAI} is an end-to-end VSR method without MEMC, as shown in Fig.\ \ref{figs1023}. The temporal information within frames is extracted by using LSTM \citep{LSTM}. Similar to RISTN \citep{RISTN}, the network consists of three parts: a spatial module, a temporal module and a reconstruction module. Spatial module is responsible for extracting features from multiple consecutive LR frames. Temporal module is a bidirectional multi-scale convoluted variant of LSTM, and is designed for extracting temporal correlation among frames.



\begin{figure}[!htbp]
\centering
\includegraphics[width=0.89\columnwidth]{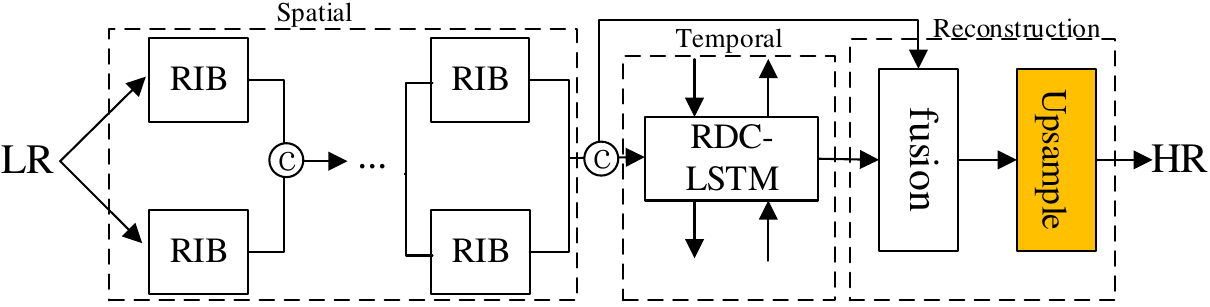}
\caption{The network architecture of RISTN \citep{RISTN}, where RIB denotes the residual invertible block, and RDC is the residual dense convolution.}
\label{figs1025}
\end{figure}

\subsubsection{RISTN}
The residual invertible spatio-temporal network (RISTN)\protect\footnote{Code: https://github.com/lizhuangzi/RISTN} \citep{RISTN} is inspired by the invertible block \citep{i-RevNet}. As shown in Fig.\ \ref{figs1025}, it designs a residual invertible block (RIB), a LSTM with residual dense convolution (RDC-LSTM), and a sparse feature fusion strategy to adaptively select useful features. Here RIB is used to extract spatial information of video frames effectively, and RDC-LSTM is used to extract spatio-temporal features.

The network is mainly divided into three parts: a spatial module, a temporal module and a reconstruction module. The spatial module is mainly composed of multiple parallel RIBs, and its output is used as the input of the temporal module. In the temporal module, after extracting spatio-temporal information, features are selectively fused by a sparse fusion strategy. Finally, the HR result of the target frame is reconstructed by the deconvolution in the reconstruction module.

\begin{figure}[!htbp]
\centering
\includegraphics[width=0.73\columnwidth]{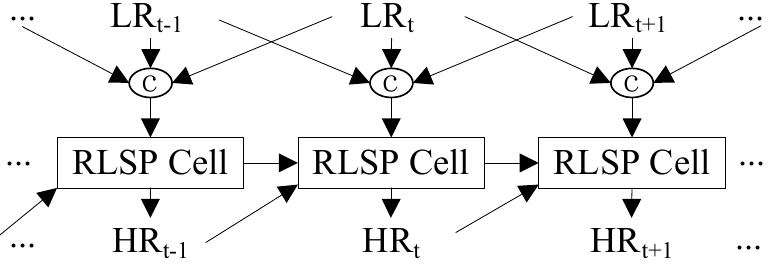}
\caption{The network architecture of RLSP \citep{fuoli2019efficient}.}
\label{figsRLSP}
\end{figure}

\subsubsection{RLSP}
The Recurrent Latent Space Propagation (RLSP)\protect\footnote{Code: https://github.com/dariofuoli/RLSP}~\citep{fuoli2019efficient} shown in Fig.\ \ref{figsRLSP} proposes a recurrent video super-resolution algorithm, which avoids the problem that a single video frame is processed multiple times in a non-recurrent network. In addition, the algorithm implicitly transmits temporal information by introducing hidden states containing the temporal information yielded in previous moment as part of the input at current moment, and does not include explicit motion estimation and motion compensation.

The hidden state is generated by the RLSP Cell, which is composed of several convolutions. The cell receives the hidden state of the previous moment, the super-resolved result of the previous moment, as well as the current frame and the adjacent frames as inputs to yield the super-resolved result and the hidden state of the current moment. This procedure repeats until all frames are processed.

\begin{figure}[!htbp]
\centering
\includegraphics[width=0.87\columnwidth]{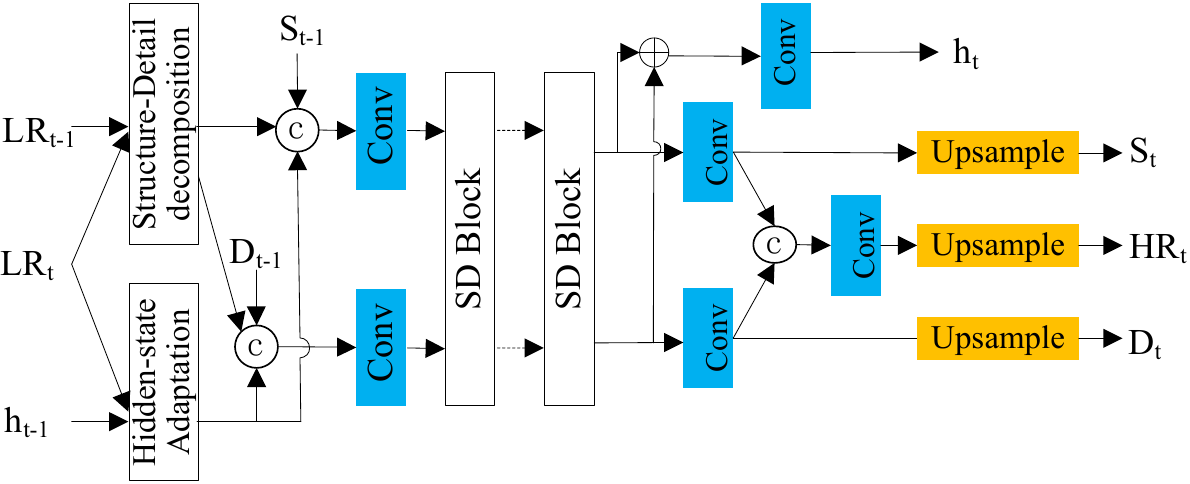}
\caption{The network architecture of RSDN \citep{isobe2020video}.}
\label{figsRSDN}
\end{figure}

\subsubsection{RSDN}
The Recurrent Structure-Detail Network (RSDN)\protect\footnote{Code: https://github.com/junpan19/RSDN}~\citep{isobe2020video} shown in Fig.\ \ref{figsRSDN} proposes to divide the frame into two components, namely structure and detail, and then process these two by subsequent module respectively.

The algorithm first uses the Bicubic interpolation algorithm to downsample and upsample the input LR frame to extract the structure and detail components. Then these two components are processed by convolution and multiple SD blocks to obtain the structure and detail components, super-resolved results and hidden states at the current moment. The SD block promotes information exchange between structure and detail components. In addition, the RSDN proposes a hidden-state adaption module to select the information that is beneficial to the super resolution and avoid the interference of redundant information.

\textbf{In summary}, the RCNN-based methods are suitable for modeling the spatio-temporal information contained in videos, since they can map neighboring frames and thus effectively establish long-term dependence with more lightweight structures. However, conventional RCNN-based methods are difficult to train and sometimes suffer from the gradient vanishing problem. And they may not capture long-term dependence when the length of input sequences is too large, and thus may not achieve great performance. LSTM-based methods can overcome these constraints to some extent with the help of the memorization of features from shallower layers. However, the complex design of LSTM is a factor that limits their depth on hardware, restraining them to model very long-term dependence.
\subsection{Non-Local Methods}
The non-local-based method is another one that utilizes both spatial and temporal information contained in video frames for super-resolution. This method benefit from the key idea of the non-local neural network \citep{NLN}, which was proposed to capture long-range dependencies for video classifications. It overcomes the flaws that convolution and recurrent computations are limited to the local area. Intuitively, a non-local operation is to calculate the response value of a position, which is equal to the weight sum of all possible positions in the input feature maps. Its formula is given as follows:
\begin{equation}
y_i=\dfrac{1}{\mathcal{C}(x)}\sum\limits_{\forall{j}}f(x_i,x_j)g(x_j)
\end{equation}
where $i$ is the index of the output location where the response value needs to be calculated, $j$ is the index of all possible locations, $x$ and $y$ are the input and output data with the same dimensions, $f$ is a function to calculate the correlation between $i$ and $j$, $g$ is the function which calculates the feature representation of input data and $\mathcal{C}(x)$ is the normalization factor. Here, $g$ is usually defined as: $g(x_j)=W_gx_j$, where $W_g$ is the weight matrix that needs to learn. It should be noted that $f$ has multiple choices such as Gaussian, dot product, and concatenation. Therefore, the non-local block can easily be added into existing deep CNNs.

\begin{figure}[!htbp]
\centering
\includegraphics[width=0.756\columnwidth]{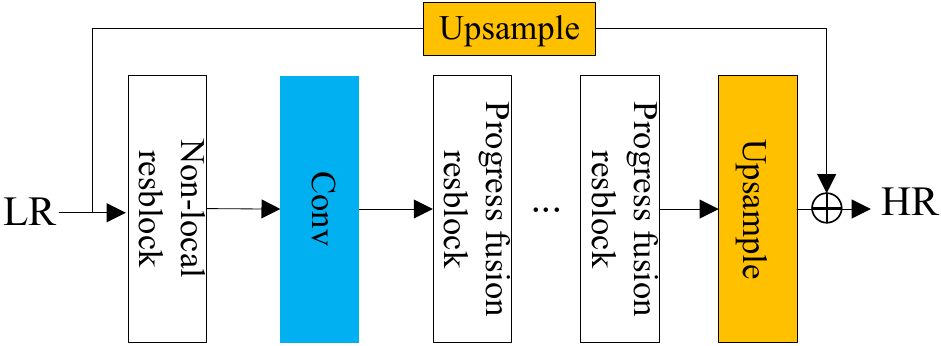}
\caption{The network architecture of PFNL \citep{PFNL}.}
\label{figs1027}
\end{figure}

\subsubsection{PFNL}
The progressive fusion non-local (PFNL) \citep{PFNL} method is illustrated in Fig.\ \ref{figs1027}. It mainly includes three parts: a non-local resblock, progressive fusion residual blocks (PFRB) and an upsampling block.

PFNL uses non-local residual blocks to extract spatio-temporal features, and PFRB is proposed to fuse them. Finally, the output through a sub-pixel convolutional layer is added to the input frame that is up-sampled by the bicubic interpolation, which is the final super-resolution result. PFRB is composed of three convolutional layers. Firstly, the input frames are convoluted with the 3$\times$3 kernels, respectively, then the output feature maps are concatenated, and the channel dimension is reduced by performing the 1$\times$1 convolution. And the results are concatenated with the previous convoluted feature maps, respectively, and conducted with a 3$\times$3 convolution. The final results are added to each input frame to obtain the output for current PFRB.
\subsection{Other}
The methods in this sub-category do not utilize the initial feature extractions mentioned above. They may combine multiple techniques for super-resolution.

\begin{figure}[!htbp]
\centering
\includegraphics[width=0.83\columnwidth]{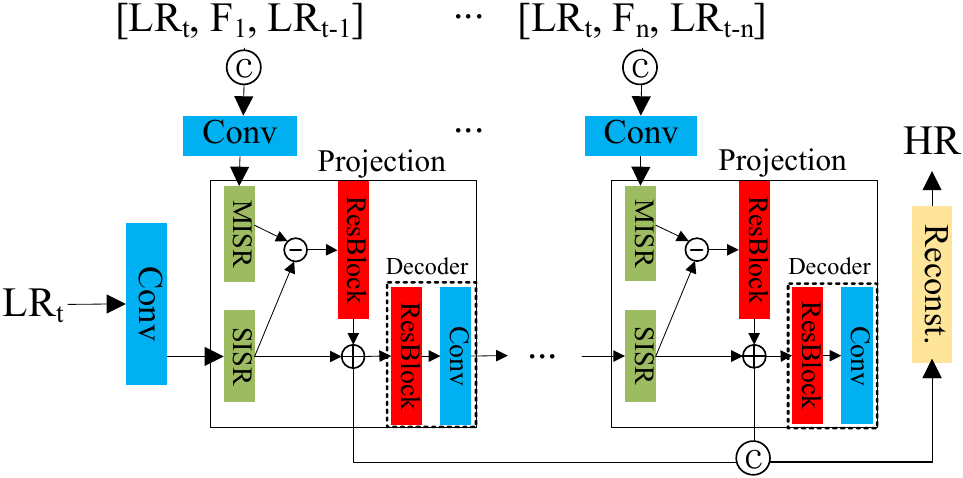}
\caption{The network architecture of RBPN~\citep{RBPN}, where $\copyright$ denotes concatenation, $\ominus$ is element subtraction, and MISR denotes multi-image super-resolution.}
\label{figs1011}
\end{figure}

\subsubsection{RBPN}
The recurrent back-projection network (RBPN)\protect\footnote{Code: https://github.com/alterzero/RBPN-PyTorch}~\citep{RBPN} is inspired by the back-projection algorithm~\citep{IRANI1991231,IRANI1993324,DBPN}. RBPN mainly consists of one feature extraction module, a projection module, and a reconstruction module, and its architecture is shown in Fig.\ \ref{figs1011}.

The feature extraction module includes two operations: One is to extract the features of the target frame, and the other is to extract the feature from the concatenation of the target frame, the neighboring frame, and the calculated optical flow which is from the neighboring frame to the target frame, and then perform alignment implicitly. The optical flow is obtained by the pyflow\protect\footnote{https://github.com/pathak22/pyflow} method. The projection module consists of an encoder and a decoder. The encoder is composed of a multiple image super-resolution (MISR), a single image super-resolution (SISR) and residual blocks (denoted as ResBlock). The decoder consists of ResBlock and a strided convolution, and it takes the output of the previous encoder as input to produce LR features for the encoder of the next projection module. The concatenation of the target frame, the next neighboring frame and pre-computed optical flow are input to the feature extraction module, whose output is also for the encoder in the next projection module. The above process does not stop until all neighboring frames are processed. That is, projection is used recurrently, which is the reason of the words ``recurrent back-projection network". Finally, the reconstruction module takes the output of the encoder in each projection module by the mean of concatenation as input to produce the final SR result.

\begin{figure}[!htbp]
\centering
\includegraphics[width=0.87\columnwidth]{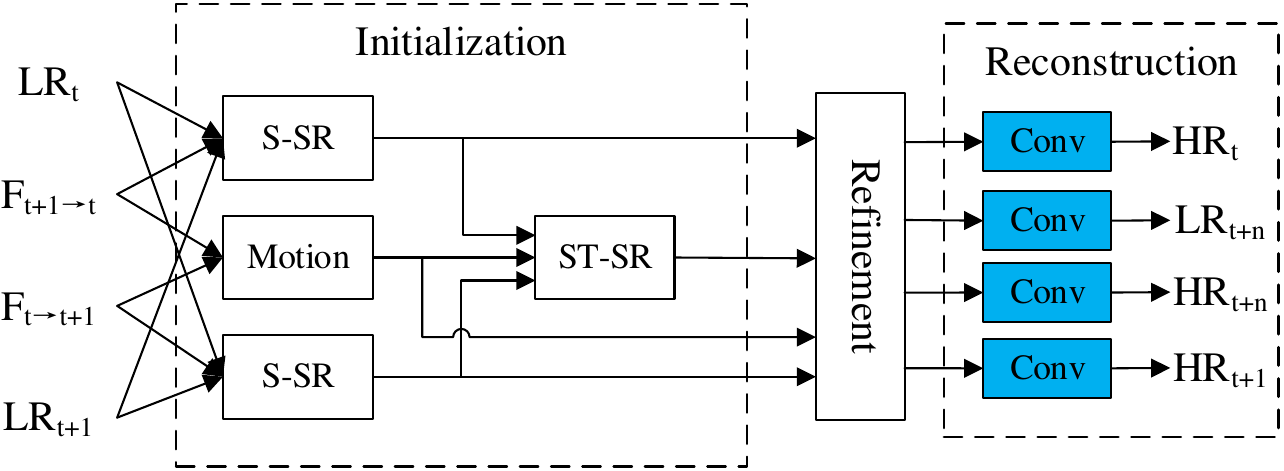}
\caption{The network architecture of STARnet \citep{STARnet}.}
\label{figstar}
\end{figure}
\subsubsection{STARnet}  
The architecture of space-time-aware multi-resolution networks (STARnet) \citep{STARnet} is shown in Fig. \ref{figstar}. STARnet is an end-to-end network that can simultaneously process video super-resolution and video interpolation. It consists of the following three stages: initialization, refinement and reconstruction.

In the initialization stage, STARnet receives four parts of inputs including two LR RGB frames and their bidirectional flow images. In this stage, the two spatial super-resolution (S-SR) modules can execute super-resolution to the two LR frames by DBPN \citep{DBPN} or RBPN \citep{RBPN} and re-generate their LR counterparts by a similar network to prepare for frame interpolation in both LR and HR spaces in the spatio-temporal super-resolution (ST-SR) module. Meanwhile, the motion module align the bidirectional flow images.

\begin{figure}[!htbp]
\centering
\includegraphics[width=0.97\columnwidth]{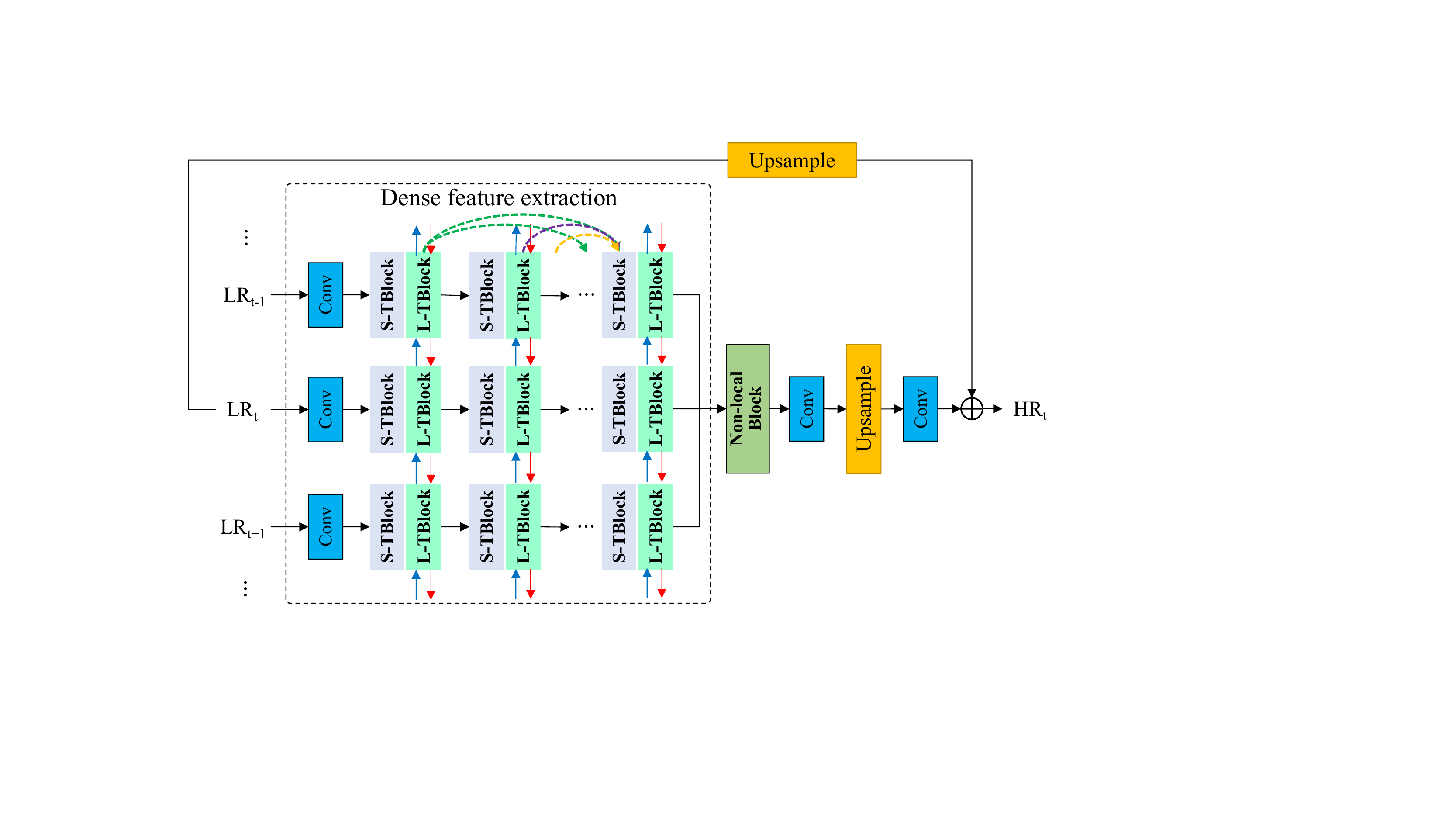}
\caption{The network architecture of DNSTNet \citep{sun2020video}.}
\label{figstar}
\end{figure}

\subsubsection{DNSTNet}
DNSTNet~\citep{sun2020video} is the video super-resolution via the Dense Non-local Spatial-Temporal convolutional Network. Here, the dense feature extraction sub-network composes of the Short-term Temporal Dependency Extraction Block (S-TBlock), Long-term TBlock (L-TBlock), and dense connections, as shown in this figure. It utilizes 3D convolution to capture short-term temporal dependency existing from adjacent frames in S-TBlock, and the bidirectional ConvLSTM for capturing long-term temporal dependency in L-TBlock. It also proposes a region-level nonlocal block following the dense feature extraction to exploit the global information, and to enlarge the limited receptive field of 3D convolution and ConvLSTM. This non-local network divides the feature maps into multiple patches and processes them respectively to decrease the computational cost. To sum up, DNSTNet adopts multiple modules to improve the performance of VSR.

Although DNSTNET uses a 3D convolution module, an LSTM module and a non-local sub-network, it does not imply better performance than EDVR and DSMC. As it is known, the network architecture is one of the important factors to affect its performance, other factors including the training strategy, and iteration number also influence its performance. Compared with the methods: EDVR and DSMC, the training strategy of DNSTNET is probably not sophisticated designed. It is a common initializing method. But EDVR is initialized by parameters from a shallower similar network. This can boost the performance. DSMC also has a deeper structure, which may be helpful to improve the performance. Moreover, in DNSTNET, too many features without selection through the dense feature concatenation are input to the non-local block for computation. These features may bring redundant information, which results in performance degradation. While in DSMC, the extracted features are refined through the U3D-RDN module before they are input to a non-local block. This processing can enhance the performance.
\begin{table*}[!htbp]
\centering
\caption{Some widely used video super-resolution datasets. Note that '*' represents unknown information.}
\resizebox{\textwidth}{!}{
\begin{tabular}{|c|c|c|m{4cm}|c|m{5cm}|c|m{2.5cm}|}
\hline
\textbf{Dataset} & \textbf{Year} & \textbf{Type} & \textbf{Download Link} & \textbf{Video Number} & \textbf{Resolution} & \textbf{Color Space} \\ \hline
        YUV25 & * & Train & \url{https://media.xiph.org/video/derf/} & 25 &$386\times 288$ & YUV \\ \hline
        TDTFF & * & Test & \url{www.wisdom.weizmann.ac.il/~vision/SingleVideoSR.html} & 5 &$648\times528 $ for Turbine, $960\times530$ for Dancing ,$700\times600$ for Treadmill, and $1000\times580$ for Flag, $990\times740$ for Fan & YUV \\ \hline
        Vid4 & 2011 & Test & \url{https://drive.google.com/drive/folders/10-gUO6zBeOpWEamrWKCtSkkUFukB9W5m} & 4 & $720\times 480$ for Foliage and Walk, $720\times 576$ for Calendar, and $704\times 576$ for City & RGB \\ \hline
        YUV21 & 2014 & Test & \url{http://www.codersvoice.com/a/webbase/video/08/152014/130.html} & 21 & $352\times 288$ & YUV \\ \hline
        Venice & 2014 & Train & \url{https://www.harmonicinc.com/free-4k-demo-footage/} & 1 & $3,840\times 2,160$ & RGB \\ \hline
        Myanmar & 2014 & Train & \url{https://www.harmonicinc.com/free-4k-demo-footage/} & 1 & $3,840\times 2,160$ & RGB \\ \hline
        CDVL & 2016 & Train & \url{http://www.cdvl.org/} & 100 & $1,920\times 1,080$ & RGB\\ \hline
        UVGD\footnotemark[21] & 2017 & Test & \url{http://ultravideo.cs.tut.fi/} & 7 & $3,840\times 2,160$ & YUV \\ \hline
	    LMT\footnotemark[22] & 2017 & Train & \url{http://mcl.usc.edu/mcl-v-database}, \url{http://live.ece.utexas.edu/research/quality/live\_video.html},\url{https://vision.in.tum.de/data/datasets}& * & $1,920\times1,080$ & RGB \\ \hline
        Vimeo-90K & 2019 & Train+Test & \url{http://toflow.csail.mit.edu/} & 91,701 & $448\times 256$ & RGB \\ \hline
        REDS\footnotemark[23] & 2019 & Train+Test & \url{https://seungjunnah.github.io/Datasets/reds.html} & 270 & $1,280\times 720$ & RGB \\ \hline
    \end{tabular}}
    \label{VSR_data}
\end{table*}
\footnotetext[21]{UVGD denotes the Ultra Video Group Database}
\footnotetext[22]{LMT denotes the LIVE video quality assessment database, the MCL-V database, and the TUM 1080p dataset.}
\footnotetext[23]{REDS denotes the REalistic and Diverse Scenes}

\begin{table*}[!htbp]
\centering
\caption{Some major video super-resolution competitions. Note that 'EDVR+' stands for a method based on EDVR, and '*' represents unknown information.}
\resizebox{\textwidth}{!}{
\begin{tabular}{|m{2.1cm}|c|c|c|m{3.2cm}|c|c|c|c|c|}
\hline
        \textbf{Name} & \textbf{Year} & \textbf{Organizer} & \textbf{Location} & \textbf{Website} & \textbf{Dataset} & \textbf{Scale} & \textbf{Champion} & \textbf{PSNR} & \textbf{SSIM} \\ \hline
        NTIRE 2019 Video Restoration and Enhancement Challenges & 2019 & CVPR & \!Long Beach, California\! & \url{https://data.vision.ee.ethz.ch/cvl/ntire19/} & REDS & $\times$4 & EDVR \citep{EDVR} & 31.79 & 0.8962 \\ \hline
        YOUKU Video Super-Resolution and Enhancement Challenge & 2019 & Alibaba & Hangzhou, China & \url{https://tianchi.aliyun.com/competition/entrance/231711/introduction} & \!Youku-VESR\! & $\times$4 & VESR-Net \citep{chen2020vesr} & 37.85 & * \\ \hline
        AIM 2019 Challenge on Video Extreme Super-Resolution & 2019 & ECCV & Hong Kong, China & \url{https://www.aim2019.org/} & Vid3oC & $\times$16 & EDVR+ & 22.53 & 0.6400 \\
        \hline
        Mobile Video Restoration Challenge & 2019 & \!ICIP \& Kwai\! & * & \url{https://www.kuaishou.com/activity/icip2019} & * & * & * &  * & * \\
        \hline
        AIM 2020 Challenge on Video Extreme Super-Resolution & 2020 & ECCV & \!Boston, Massachusetts\! & \url{http://aim2020.org/} & Vid3oC & $\times$16 & \!EVESRNet \citep{EVESRNet}\! & 22.83 & 0.6450 \\ \hline
        Mobile AI 2021 Real-Time Video Super-Resolution Challenge & 2021 & CVPR & \!  VIRTUAL  \! & \url{https://ai-benchmark.com/workshops/mai/2021/} & REDS & $\times$4 & Diggers~\citep{ignatov2021realvideo}  & 28.33 & 0.8112  \\ \hline
        NTIRE  2021 Video Super-Resolution Challenge & 2021 & CVPR & \!  VIRTUAL  \! & \url{https://data.vision.ee.ethz.ch/cvl/ntire21/} & REDS & $\times$4 & BasicVSR++ \citep{BasicVSR++}  & 33.36 & 0.9218  \\
        \hline
    \end{tabular}}
    \label{VSR_races}
\end{table*}

\textbf{In summary}, the non-local based methods introduce attention mechanisms into VSR tasks. They can establish effective dependence of spatio-temporal information by extending the receptive field to the global. However, the non-local modules used in them need to calculate the response at each position by attending to all other positions and computing a weighted average of the features in all positions. Thus, this incurs high computational cost, and some efforts can be made to reduce the computational overhead of the methods.

Moreover, the methods without alignment rely on the non-linear capability of the neural network to learn the motion correlation between frames for video super-resolution. They do not utilize additional modules to align frames. The learning ability largely depends on the design of the deep neural network. And an elaborate design is more likely leading to higher performance for video super-resolution.

In addition, we discuss the deeper connections between all the methods below.
1)	The methods such as EDVR, DNLN, TDAN, D3Dnet and VESR-Net, which belong to the deformable convolution category, all attempt to overcome the flaw of optical flow-based methods by using the DConv structure. The estimation of optical flow is inaccurate when dealing with complex motions and varying illumination, while the receptive field of the convolution kernel can be expanded by utilizing DConv. And the network can better capture complex motions and illumination changes.
2)	The methods such as DUF, FSTRN, 3DSRnet and DSMC all employ 3D convolutional layers to learn spatial and temporal features simultaneously instead of the 2D convolution from video data. Besides, they also try to avoid the inaccuracy of motion estimation and compensation when complex motions involve by designing new network structures.
3)	The methods such as BRCN, STCN, RISTN, RLSP, RSDN and BasicVSR exploit long-term contextual information contained in video frames by using bidirectional recurrent convolutional networks. The bidirectional RCNN can utilize temporal dependency from both previous and future frames through the combination of a forward recurrent network and a backward recurrent network.
4)	The methods such as RVSR, STCN, BRCN, EDVR, DNLN, TDAN, D3DNet, VESR-Net, DUF, 3DSRNet and DSMC involve the dealing with complex motions in the videos.
5)	The methods such as MuCAN (in MEMC class), EDVR (in DC class), VESR-Net (in DC class) and PFNL (in non-local class) attempt to capture the global dependency between different positions across frame. Specifically, the TSA module in the EDVR method assigns pixel-level weights on each frame for fusion. MuCAN, VESR-Net, and PFNL all design non-local modules to correlate different patches, which improve the ability to capture motion information.
6)	The methods such as DRVSR, MultiBoot VSR, and DSMC all address the video super-resolution with multiple scaling factors. They not only consider x4 scale, but also regard $\times$2, $\times$3, or $\times$8 scales.
7)	The methods such as MultiBoot VSR, PFNL and RBPN all pay attention to improve the training strategies. For example, PFNL adopts residual learning to stabilize the training process.

\section{Performance Comparisons}
\subsection{Datasets and Competitions}
Details of some of the most popular datasets used in VSR tasks are summarized in Table \ref{VSR_data}. The most widely-used dataset for training is Vimeo-90K, since it is currently the largest VSR dataset with real scenes. The most popular dataset for testing is Vid4, whose frames contain more high-frequency details than others. Thus, Vid4 is frequently used for evaluating the performance of VSR methods. REDS includes videos with extremely large movement, which is challenging for VSR methods.

Besides, we also summarize several international competitions on video super-resolution in Table~\ref{VSR_races}. The NTIRE 2019 Challenge~\citep{Nah_2019_CVPR_Workshops1,Nah_2019_CVPR_Workshops2} aims at recovering videos with large movements and diverse real-world scenes. Its winning solution is EDVR~\citep{EDVR}, which may be one of the most popular works for VSR. The AIM Challenges in 2019~\citep{fuoli2019aim} and 2020~\citep{fuoli2020aim} both encourage solutions of VSR with large scale factors. A method enhanced from EDVR won the AIM 2019 Challenge, while EVESRNet~\citep{EVESRNet} won the AIM 2020 Challenge. Besides, the YOUKU Video Super-Resolution and Enhancement Challenge, and Mobile Video Restoration Challenge in 2019 are both for videos which are more relevant to entertainment. The winning solution of YOUKU challenge is VESR-Net~ \citep{chen2020vesr}. The Mobile AI 2021 Real-Time Video Super-Resolution challenge~\citep{ignatov2021realvideo} evaluated the solutions on an OPPO Find X2 smartphone GPU. The most recent NTIRE 2021 Challenge on Video Super-Resolution gauges the state-of-the-art~\citep{son2021ntire}, its winner being BasicVSR++~\citep{BasicVSR++}. These competitions are making great contributions to the development of video super-resolution and helping develop new methods for various video super-resolution applications.

\begin{table*}[!h]
	\caption{Comparison of all the methods on the datasets with scale factor $\times4$. Note that `Internet' means that the dataset is collected from the internet. `*' denotes that the source of the dataset is unknown, and `-' indicates that the method does not be tested on the datasets. }
	\centering
	\vspace{-2mm}	
	\begin{threeparttable}
		\resizebox{\textwidth}{!}{
			\begin{tabular}{|c|c|c|c|c|cc|cc|}
				\hline
				\multirow{2}{*}{Method}   & \multirow{2}{*}{Training Set}                                 & \multirow{2}{*}{Test Set} & \multirow{2}{*}{Channel} & \multirow{2}{*}{Params.(MB)} & \multicolumn{2}{c|}{BI}             & \multicolumn{2}{c|}{BD}             \\ \cline{6-9}
				&                                                               &                           &                          &                         & \multicolumn{1}{c|}{PSNR}  & SSIM   & \multicolumn{1}{c|}{PSNR}  & SSIM   \\ \hline
				Deep-DE                   & *                                                             & city+temple+penguin       & *                        & $ 1.11^{[1]} $          & \multicolumn{1}{c|}{-}     & -      & \multicolumn{1}{c|}{29.00} & 0.8870 \\ \hline
				\multirow{2}{*}{VSRnet}   & \multirow{2}{*}{Myanmar}                                      & Vid4                      & Y                        & \multirow{2}{*}{$ 0.27^{[2]} $}   & \multicolumn{1}{c|}{24.84} & 0.7049 & \multicolumn{1}{c|}{-}     & -      \\ \cline{3-4} \cline{6-9}
				&                                                               & Myanmar--T                & Y                        &                         & \multicolumn{1}{c|}{31.85} & 0.8834 & \multicolumn{1}{c|}{-}     & -      \\ \hline
				VESPCN                    & CDVL                                                          & Vid4                      & Y                        & $ 0.88^{[2]} $          & \multicolumn{1}{c|}{25.35} & 0.7557 & \multicolumn{1}{c|}{-}     & -      \\ \hline
				\multirow{2}{*}{DRVSR}    & \multirow{2}{*}{*}                                            & Vid4                      & Y                        & \multirow{2}{*}{$ 2.17^{[3]} $}   & \multicolumn{1}{c|}{25.52} & 0.7600 & \multicolumn{1}{c|}{-}     & -      \\ \cline{3-4} \cline{6-9}
				&                                                               & SPMCS                     & Y                        &                         & \multicolumn{1}{c|}{29.69} & 0.8400 & \multicolumn{1}{c|}{-}     & -      \\ \hline
				\multirow{2}{*}{RVSR}     & \multirow{2}{*}{LMT}                                          & Vid4+temple+penguin       & Y                        & \multirow{2}{*}{-}      & \multicolumn{1}{c|}{28.05} & -      & \multicolumn{1}{c|}{-}     & -      \\ \cline{3-4} \cline{6-9}
				&                                                               & UVGD                      & Y                        &                         & \multicolumn{1}{c|}{39.71} & -      & \multicolumn{1}{c|}{-}     & -      \\ \hline
				\multirow{2}{*}{FRVSR}    & \multirow{2}{*}{Vimeo-90K}                                    & Vid4                      & Y                        & \multirow{2}{*}{$ 2.81^{[3]} $}   & \multicolumn{1}{c|}{-}     & -      & \multicolumn{1}{c|}{26.69} & 0.8103 \\ \cline{3-4} \cline{6-9}
				&                                                               & Vimeo-90K-T               & Y                        &                         & \multicolumn{1}{c|}{-}     & -      & \multicolumn{1}{c|}{35.64} & 0.9319 \\ \hline
				\multirow{2}{*}{SOFVSR}   & \multirow{2}{*}{CDVL}                                         & DAVIS-10                  & Y                        & \multirow{2}{*}{$ 1.71^{[3]} $}   & \multicolumn{1}{c|}{34.32} & 0.9250 & \multicolumn{1}{c|}{34.27} & 0.9250 \\ \cline{3-4} \cline{6-9}
				&                                                               & Vid4                      & Y                        &                         & \multicolumn{1}{c|}{26.01} & 0.7710 & \multicolumn{1}{c|}{26.19} & 0.7850 \\ \hline
				\multirow{2}{*}{TecoGAN}  & \multirow{2}{*}{*}                                            & ToS                       & Y                        & \multirow{2}{*}{3.00}   & \multicolumn{1}{c|}{-}     & -      & \multicolumn{1}{c|}{32.75} & -      \\ \cline{3-4} \cline{6-9}
				&                                                               & Vid4                      & Y                        &                         & \multicolumn{1}{c|}{-}     & -      & \multicolumn{1}{c|}{25.89} & -      \\ \hline
				\multirow{2}{*}{TOFlow}   & \multirow{2}{*}{Vimeo-90K}                                    & Vid4                      & Y                        & \multirow{2}{*}{$ 1.41$}   & \multicolumn{1}{c|}{23.54} & 0.8070 & \multicolumn{1}{c|}{-}     & -      \\ \cline{3-4} \cline{6-9}
				&                                                               & Vimeo-90K-T               & Y                        &                         & \multicolumn{1}{c|}{33.08} & 0.9417 & \multicolumn{1}{c|}{-}     & -      \\ \hline
				\multirow{4}{*}{MMCNN}    & \multirow{4}{*}{*}                                            & Vid4                      & Y                        & \multirow{4}{*}{10.58}  & \multicolumn{1}{c|}{26.28} & 0.7844 & \multicolumn{1}{c|}{-}     & -      \\ \cline{3-4} \cline{6-9}
				&                                                               & Myanmar-T                 & Y                        &                         & \multicolumn{1}{c|}{33.06} & 0.9040 & \multicolumn{1}{c|}{-}     & -      \\ \cline{3-4} \cline{6-9}
				&                                                               & YUV21                     & Y                        &                         & \multicolumn{1}{c|}{28.90} & 0.7983 & \multicolumn{1}{c|}{-}     & -      \\ \cline{3-4} \cline{6-9}
				&                                                               & Vid4+temple+penguin       & Y                        &                         & \multicolumn{1}{c|}{28.97} & -      & \multicolumn{1}{c|}{-}     & -      \\ \hline
				\multirow{2}{*}{MEMC-Net} & \multirow{2}{*}{Vimeo-90K}                                    & Vimeo-90K-T               & Y                        & \multirow{2}{*}{-}      & \multicolumn{1}{c|}{33.47} & 0.9470 & \multicolumn{1}{c|}{-}     & -      \\ \cline{3-4} \cline{6-9}
				&                                                               & Vid4                      & Y                        &                         & \multicolumn{1}{c|}{24.37} & 0.8380 & \multicolumn{1}{c|}{-}     & -      \\ \hline
				\multirow{3}{*}{RRCN}     & \multirow{3}{*}{Myanmar}                                      & Myanmar-T                 & Y                        & \multirow{3}{*}{-}      & \multicolumn{1}{c|}{32.35} & 0.9023 & \multicolumn{1}{c|}{-}     & -      \\ \cline{3-4} \cline{6-9}
				&                                                               & Vid4                      & Y                        &                         & \multicolumn{1}{c|}{25.86} & 0.7591 & \multicolumn{1}{c|}{-}     & -      \\ \cline{3-4} \cline{6-9}
				&                                                               & YUV21                     & Y                        &                         & \multicolumn{1}{c|}{29.08} & 0.7986 & \multicolumn{1}{c|}{-}     & -      \\ \hline
				\multirow{2}{*}{RTVSR}    & \multirow{2}{*}{harmonicinc.com}                              & Vid4                      & Y                        & \multirow{2}{*}{15.00}  & \multicolumn{1}{c|}{26.36} & 0.7900 & \multicolumn{1}{c|}{-}     & -      \\ \cline{3-4} \cline{6-9}
				&                                                               & Vid4+temple+penguin       & Y                        &                         & \multicolumn{1}{c|}{29.03} & -      & \multicolumn{1}{c|}{-}     & -      \\ \hline
				MultiBoot VSR                & REDS                                                          & REDS-T                    & RGB                      & 60.86                   & \multicolumn{1}{c|}{31.00} & 0.8822 & \multicolumn{1}{c|}{-}     & -      \\ \hline
				\multirow{2}{*}{MuCAN}    & Vimeo-90K                                                     & Vimeo-90K-T               & Y                        & 19.90                      & \multicolumn{1}{c|}{37.32} & 0.9465 & \multicolumn{1}{c|}{-}     & -      \\ \cline{2-4} \cline{5-9}
				&                           REDS                                                          & REDS4                     & RGB                      & 25.70                   & \multicolumn{1}{c|}{30.88} & 0.8750 & \multicolumn{1}{c|}{-}     & -      \\ \hline
				\multirow{4}{*}{IconVSR}  & \multirow{3}{*}{Vimeo-90K}                                    & Vimeo-90K-T               & Y                        & \multirow{4}{*}{8.70}   & \multicolumn{1}{c|}{37.47} & 0.9476 & \multicolumn{1}{c|}{37.84} & 0.9524 \\ \cline{3-4} \cline{6-9}
				&                                                               & Vid4                      & Y                        &                         & \multicolumn{1}{c|}{27.39} & 0.8279 & \multicolumn{1}{c|}{28.04} & 0.8570 \\ \cline{3-4} \cline{6-9}
				&                                                               & UDM10                     & Y                        &                         & \multicolumn{1}{c|}{-}     & -      & \multicolumn{1}{c|}{40.03} & 0.9694 \\ \cline{2-4} \cline{6-9}
				&                           REDS                                                          & REDS4                     & RGB                      &                         & \multicolumn{1}{c|}{31.67} & 0.8948 & \multicolumn{1}{c|}{-}     & -      \\ \hline
				\multirow{3}{*}{EDVR}     & \multirow{2}{*}{Vimeo-90K}                                    & Vid4                      & Y                        & \multirow{3}{*}{$ 20.60$}  & \multicolumn{1}{c|}{27.35} & 0.8264 & \multicolumn{1}{c|}{27.85} & 0.8503 \\ \cline{3-4} \cline{6-9}
				&                                                               & Vimeo-90K-T               & Y                        &                         & \multicolumn{1}{c|}{37.61} & 0.9489 & \multicolumn{1}{c|}{37.81} & 0.9523 \\ \cline{2-4} \cline{6-9}
				&                           REDS                                                          & REDS4                     & RGB                      &                         & \multicolumn{1}{c|}{31.09} & 0.8800 & \multicolumn{1}{c|}{28.88} & 0.8361 \\ \hline
				\multirow{2}{*}{DNLN}     & \multirow{2}{*}{Vimeo-90K}                                    & Vid4                      & Y                        & \multirow{2}{*}{19.74}  & \multicolumn{1}{c|}{27.31} & 0.8257 & \multicolumn{1}{c|}{-}     & -      \\ \cline{3-4} \cline{6-9}
				&                                                               & SPMCS                     & Y                        &                         & \multicolumn{1}{c|}{30.36} & 0.8794 & \multicolumn{1}{c|}{-}     & -      \\ \hline
				TDAN                      & Vimeo-90K                                                     & Vid4                      & Y                        & $ 1.97^{[2]} $          & \multicolumn{1}{c|}{26.24} & 0.7800 & \multicolumn{1}{c|}{26.58} & 0.8010 \\ \hline
				D3Dnet                    & Vimeo-90K                                                     & Vid4                      & Y                        & 2.58                    & \multicolumn{1}{c|}{26.52} & 0.7990 & \multicolumn{1}{c|}{-}     & -      \\ \hline
				VESR-Net                  & Youku-VESR                                                    & Youku-VESR-T              & RGB                      & 21.65                   & \multicolumn{1}{c|}{-}     & -      & \multicolumn{1}{c|}{35.97} & -      \\ \hline
				VSRResFeatGAN             & Myanmar                                                       & Vid4                      & Y                        & -                       & \multicolumn{1}{c|}{25.51} & 0.7530 & \multicolumn{1}{c|}{-}     & -      \\ \hline
				FFCVSR                    & Venice+Myanmar                                                & Vid4                      & Y                        & -                       & \multicolumn{1}{c|}{26.97} & 0.8300 & \multicolumn{1}{c|}{-}     & -      \\ \hline
				\multirow{3}{*}{DUF}      & \multirow{2}{*}{Vimeo-90K}                                    & Vid4                      & Y                        & \multirow{3}{*}{$ 5.82 $}   & \multicolumn{1}{c|}{-}     & -      & \multicolumn{1}{c|}{27.38} & 0.8329 \\ \cline{3-4} \cline{6-9}
				&                                                               & Vimeo-90K-T               & Y                        &                         & \multicolumn{1}{c|}{-}     & -      & \multicolumn{1}{c|}{36.87} & 0.9447 \\ \cline{2-4} \cline{6-9}
				&                           REDS                                                          & REDS4                     & Y                        &                         & \multicolumn{1}{c|}{28.63} & 0.8251 & \multicolumn{1}{c|}{-}     & -      \\ \hline
				FSTRN                     & YUV25                                                         & TDTFF                     & Y                        & -                       & \multicolumn{1}{c|}{-}     & -      & \multicolumn{1}{c|}{29.95} & 0.8700 \\ \hline
				3DSRnet                   & largeSet                                                      & Vid4                      & Y                        & $ 0.11^{[3]} $          & \multicolumn{1}{c|}{25.71} & 0.7588 & \multicolumn{1}{c|}{-}     & -      \\ \hline
				\multirow{2}{*}{DSMC}     & \multirow{2}{*}{REDS}                                         & REDS4                     & RGB                      & \multirow{2}{*}{11.58}  & \multicolumn{1}{c|}{30.29} & 0.8381 & \multicolumn{1}{c|}{-}     & -      \\ \cline{3-4} \cline{6-9}
				&                                                               & Vid4                      & Y                        &                         & \multicolumn{1}{c|}{27.29} & 0.8403 & \multicolumn{1}{c|}{-}     & -      \\ \hline
				\multirow{2}{*}{BRCN}     & \multirow{2}{*}{YUV25}                                        & Vid4                      & Y                        & \multirow{2}{*}{-}      & \multicolumn{1}{c|}{-}     & -      & \multicolumn{1}{c|}{24.43} & 0.6334 \\ \cline{3-4} \cline{6-9}
				&                                                               & TDTFF                     & Y                        &                         & \multicolumn{1}{c|}{-}     & -      & \multicolumn{1}{c|}{28.20} & 0.7739 \\ \hline
				\multirow{2}{*}{STCN}     & \multirow{2}{*}{*}                                            & Hollywood2                & Y                        & \multirow{2}{*}{-}      & \multicolumn{1}{c|}{-}     & -      & \multicolumn{1}{c|}{34.58} & 0.9259 \\ \cline{3-4} \cline{6-9}
				&                                                               & city+temple+penguin       & *                        &                         & \multicolumn{1}{c|}{-}     & -      & \multicolumn{1}{c|}{30.27} & 0.9103 \\ \hline
				RISTN                     & \begin{tabular}[c]{@{}c@{}} Vimeo-90K \end{tabular} & Vid4                      & Y                        & 3.67                    & \multicolumn{1}{c|}{26.13} & 0.7920 & \multicolumn{1}{c|}{-}     & -      \\ \hline
				\multirow{2}{*}{RLSP}     & \multirow{2}{*}{Vimeo-90K}                                    & Vid4                      & Y                        & \multirow{2}{*}{$ 4.21$}   & \multicolumn{1}{c|}{-}     & -      & \multicolumn{1}{c|}{27.48} & 0.8388 \\ \cline{3-4} \cline{6-9}
				&                                                               & Vimeo-90K-T               & Y                        &                         & \multicolumn{1}{c|}{-}     & -      & \multicolumn{1}{c|}{36.49} & 0.9403 \\ \hline
				\multirow{3}{*}{RSDN}     & \multirow{3}{*}{Vimeo-90K}                                    & Vid4                      & Y                        & \multirow{3}{*}{6.19}   & \multicolumn{1}{c|}{-}     & -      & \multicolumn{1}{c|}{27.92} & 0.8505 \\ \cline{3-4} \cline{6-9}
				&                                                               & Vimeo-90K-T               & Y                        &                         & \multicolumn{1}{c|}{-}     & -      & \multicolumn{1}{c|}{37.23} & 0.9471 \\ \cline{3-4} \cline{6-9}
				&                                                               & UDM10                     & Y                        &                         & \multicolumn{1}{c|}{-}     & -      & \multicolumn{1}{c|}{39.35} & 0.9653 \\ \hline
				\multirow{3}{*}{PFNL}     & \multirow{2}{*}{Vimeo-90K}                                    & Vid4                      & Y                        & \multirow{3}{*}{$ 3.00 $}   & \multicolumn{1}{c|}{26.73} & 0.8029 & \multicolumn{1}{c|}{27.16} & 0.8355 \\ \cline{3-4} \cline{6-9}
				&                                                               & Vimeo-90K-T               & Y                        &                         & \multicolumn{1}{c|}{36.14} & 0.9363 & \multicolumn{1}{c|}{-}     & -      \\ \cline{2-4} \cline{6-9}
				&                           REDS                                                          & REDS4                     & RGB                      &                         & \multicolumn{1}{c|}{29.63} & 0.8502 & \multicolumn{1}{c|}{-}     & -      \\ \hline
				\multirow{3}{*}{RBPN}     & \multirow{2}{*}{Vimeo-90K}                                    & Vid4                      & Y                        & \multirow{3}{*}{$ 12.20 $}  & \multicolumn{1}{c|}{27.12} & 0.8180 & \multicolumn{1}{c|}{-}     & -      \\ \cline{3-4} \cline{6-9}
				&                                                               & Vimeo-90K-T               & Y                        &                         & \multicolumn{1}{c|}{37.07} & 0.9453 & \multicolumn{1}{c|}{37.20} & 0.9458 \\ \cline{2-4} \cline{6-9}
				&                           REDS                                                          & REDS4                     & RGB                      &                         & \multicolumn{1}{c|}{30.09} & 0.8590 & \multicolumn{1}{c|}{-}     & -      \\ \hline
				\multirow{3}{*}{STARnet}  & \multirow{3}{*}{Vimeo-90K}                                    & UCF101                    & *                        & \multirow{3}{*}{$ 111.61^{[4]} $} & \multicolumn{1}{c|}{29.11} & 0.9240 & \multicolumn{1}{c|}{-}     & -      \\ \cline{3-4} \cline{6-9}
				&                                                               & Vimeo-90K-T               & *                        &                         & \multicolumn{1}{c|}{30.83} & 0.9290 & \multicolumn{1}{c|}{-}     & -      \\ \cline{3-4} \cline{6-9}
				&                                                               & Middlebury                & *                        &                         & \multicolumn{1}{c|}{27.16} & 0.8270 & \multicolumn{1}{c|}{-}     & -      \\ \hline
				\multirow{3}{*}{DNSTNet}  & \multirow{3}{*}{Vimeo-90K}                                    & Vid4                      & Y                        & \multirow{3}{*}{-}      & \multicolumn{1}{c|}{27.21} & 0.8220 & \multicolumn{1}{c|}{-}     & -      \\ \cline{3-4} \cline{6-9}
				&                                                               & Vimeo-90K-T               & Y                        &                         & \multicolumn{1}{c|}{36.86} & 0.9387 & \multicolumn{1}{c|}{-}     & -      \\ \cline{3-4} \cline{6-9}
				&                                                               & SPMCS                     & Y                        &                         & \multicolumn{1}{c|}{29.74} & 0.8710 & \multicolumn{1}{c|}{-}     & -      \\ \hline
		\end{tabular}}
	\end{threeparttable}
\label{tab:Margin_settings_2}
\end{table*}

\subsection{Performance of Methods}
Moreover, we summarize the performance of the representative VSR methods with scale factor $4$ in Table \ref{tab:Margin_settings_2} in terms of both PSNR and SSIM. More experimental results for VSR tasks with magnification factors 2 and 3 are reported in Supplementary Materials. The degradation types are the bicubic downsampling with the image-resize function (BI) and Gaussian blurring and downsampling (BD). Note that part of the PSNR and SSIM are from their original works. And a simple comparison on the performance may not be fair, since the training data, the pre-processing, and the cropped area in videos are likely totally different in the methods. The details about the performance are listed to provide reference for readers.



According to Table \ref{tab:Margin_settings_2}, the top 5 methods in the $\times 4$ VSR task on Vimeo-90K-T dataset are as follows. The methods are denoted by (PSNR, BI/BD, Params.).
IconVSR (37.84, BD, 8.70), EDVR (37.61, BI, 20.60), IconVSR (37.47, BI, 8.70), MuCAN (37.32, BI, 19.90), and RSDN (37.23, BD, 6.19). The top 5 methods on Vid4 dataset are IconVSR (28.04, BD, 8.70), RSDN (27.92, BD, 6.19), EDVR (27.85, BD, 20.60), RLSP (27.48, BD, 4.21), and DUF (27.38, BD, 5.82). The top 4 methods on REDS4 dataset are IconVSR (31.67, BI, 8.70), EDVR (31.09, BI, 20.60), MuCAN (30.88, BI, 25.70), and DSMC (30.29, BI, 11.58). In the method evaluation, we compare the results on Y channel for Vimeo-90K-T and Vid4 datasets, and on RGB channel for REDS4. PFNL and DNLN do not utilize all the test frames.

IconVSR and EDVR show superior performance on the three datasets. And IconVSR uses optical flow for feature alignment, a bidirectional recurrent network for temporal feature propagation, and an information-refill mechanism for feature refinement. With these properties, it outperforms some other methods in some cases, and achieves more performance gain with BD degradation than BI degradation on Vimeo-90K-T and Vid4. EDVR employs cascaded multi-scale deformable convolutions for alignment, and TSA to fuse multiple frames. Unlike DNLN, which also adopts deformable convolutions, EDVR can capture multi-scale feature information. Compared with TDAN and D3Dnet, the architecture of EDVR is more complicated and may learn more information from inputs, though they all employ deformable convolutions for alignment. And EDVR costs 20.60 MB parameters, which is far more than other top networks. This may explain its better performance.

For the Vid4 dataset, RLSP and RSDN both adopt recurrent convolutional neural network as backbone to utilize the temporal information contained in multiple frames. RSDN further divides a frame into structure and detail to process them respectively, and also exchange the information between them. This refined extraction attributes to its performance. PFNL proposes the non-local residual block to capture long-range spatio-temporal dependencies between frames, which may outperform some conventional MEMC-based methods.

For the Vimeo-90K-T dataset, the performance of MuCAN is likely attributed to the two main modules, CN-CAM and TM-CAM. The former module can hierarchically aggregate information for handling large and subtle motion, and the latter one captures nonlocal communication within different feature resolutions. RSDN relies the information exchange between the structure and detail to gain better performance on Vimeo-90K-T. It is noticed that MuCAN has 19.90 MB parameters, which is far more than those of RSDN and IconVSR on this dataset.

Moreover, for the REDS4 dataset, it is noticed that EDVR and MuCAN both have more than 20.0 MB parameters, which is far more than those of IconVSR and DSMC, though they are in the second and the third places on the top list. DSMC proposes the U3D-RDN module which learns coarse-to-fine spatio-temporal features, and MSCU, which decomposes an upsampling into multiple sub-tasks for full use of the intermediate results as well as a dual subnet for aiding the training. DSMC demonstrates superior performance to other 3D convolution methods.

\subsection{Guidelines for Model Selection}
In this subsection, we provide some guidelines for readers to select different models according to the results in Table \ref{tab:Margin_settings_2}. For the super-resolving videos with realistic textures and rich details but without large motions, the following methods can be prime candidates: IconVSR, RSDN, EDVR, RLSP, DUF, DNLN, DSMC, PFNL, RBPN, and FRVSR. These methods are ordered according to the PSNR values on the Vid4 dataset, whose videos contain more high-frequency details. Among them, EDVR and DNLN both have more than 20.0 MB parameters, which are suitable for the applications without tight restriction on GPU memory. And the methods such as IconVSR, RSDN, RLSP, DUF, PFNL and FRVSR cost less than 10.0 MB model parameters, which might be more appropriate for the application of mobile devices and embedding systems.

When dealing with video sequences with complex and large motions, one can select the methods, IconVSR, EDVR, DSMC, RBPN, and PFNL. The performance of these methods is ranged in descending order and referred to their PSNR results on the REDS dataset. Similar to the above applications, the number of the parameters in EDVR exceeds 20.0 MB, while those of IconVSR and PFNL are fewer than 10.0 MB.

For generic videos except for the above two videos, we recommend the methods, IconVSR, EDVR, MuCAN, RSDN, RBPN, RLSP, PFNL and FRVSR. These methods are ordered according to the PSNR values on the Vimeo-90k-T dataset. The number of the parameters in EDVR is larger than 20.0 MB, and those of IconVSR, MuCAN, RSDN, RLSP, PFNL and FRVSR are fewer than 10.0 MB.

There are some additional tips for selecting the methods with alignment. When inaccurate motion estimation and alignment may introduce artifacts for videos with large motions or lighting changes, the deformable convolution-based methods are more robust for VSR tasks. When considering the online applications of video super-resolution, a unidirectional network may be the best candidate, where the information is sequentially propagated from the first frame to the last frame. While for offline applications, a bidirectional network in which the features can propagate forward and backward in time independently, is a better choice for VSR. In this case, the optical flow can be estimated both sequentially and reversely. It is known that the motion estimation is one critical step for the methods with alignment, which directly influences the performance of VSR methods. When more advanced estimation methods are proposed, they can be used to improve VSR's performance.

\section{Applications of Video Super-Resolution}
By using VSR techniques, the resolution of video frames can be enhanced, and better visual quality and recognition accuracy can be achieved. It has a variety of applications, such as remote sensing, medical diagnoses, video decoding, and 3D reconstruction.

\subsection{Video Decoding}
In \citep{glaister2011hybrid}, a patch-based super-resolution method was presented to decode frames for video playback, and had been integrated in a video compression pipeline. \citet{dai2015dictionary} proposed a VSR algorithm based on dictionary learning and sub-pixel motion compensation. This algorithm adopted multiple bilevel dictionaries for single-frame SR. Meanwhile, they presented a dictionary learning algorithm, where the dictionaries are trained from consecutive video frames. In \citep{liu2018new}, an improved super-resolution reconstruction algorithm, which was a part of the proposed low bit-rate coding scheme, was applied to the decoded data for reconstructing high-definition videos. In \citep{umeda2018hdr}, an anchored neighborhood regression SR method \citep{timofte2014a+} was used for decoding in the proposed video coding system.

 \citet{kim20182x} proposed a hardware-friendly VSR algorithm which can upscale full-high-definition (FHD) video streams to their 4K ultra-high-definition counterparts, and implemented it in both field programmable gate array (FPGA) and application specific integrated circuit (ASIC) hardware for real-time video reconstruction. They further presented a FPGA-based network structure for SR. The number of parameters is reduced by using cascaded convolutions and depth-wise separable residual network \citep{kim2018real}. In \citep{wei2019fpga}, a CNN-based SR algorithm was implemented and accelerated through network pruning and quantization, and the algorithm was integrated in their real-time FPGA-based system, which supports video stream transcoding from H.264 FHD to H.265/HEVC UHD.

\subsection{Remote Sensing}
The image SR methods such as VDSR and ESPCN have been utilized to enhance the resolution of objects in satellite videos in \citep{luo2017video, xiao2018super}. In \citep{jiang2018progressively}, a progressively enhanced network with a transition unit was proposed to strengthen residual images with fine details. Moreover, \citet{jiang2018deep} proposed a deep distillation recursive network with a multi-scale purification unit to super-resolve the images in the Jilin-1 satellite videos. \citet{liu2020satellite} proposed a framework to pose the image priors in maximum a posteriori to regularize the solution space and generate the corresponding high-resolution video frames. The framework combines the implicitly captured local motion information through exploiting spatiotemporal neighbors and the nonlocal spatial similarity to recover HR frames. The experiments on the videos from the Jilin-1 satellite and the OVS-1A satellite verify that the approach can preserve edges and texture details.

\subsection{Medical Analysis}
\citet{poot2010general} and \citet{odille2015motion} reconstructed isotropic 3D magnetic resonance imaging (MRI) data in high resolution from multiple low-resolution MRI slices of different orientations, and they did not utilize accurate motion estimation and alignment. In \citep{zhang2012reconstruction}, HR 4D computerized tomography (CT) images are super-resolved with several frames for each slice at different respiratory phases. \citet{yu2017computed} proposed a multi-slice CT SR network, which inputs the consecutive CT slices as video frames. It is composed of several convolution layers and a rearranging layer, and a subset of 5,800 slices is used to train the model, and the other 1,000 slices for testing.

\citet{ren2019towards} proposed a framework, which adopts an iterative upsampling layer and one downsampling layer in DBPN \citep{DBPN} to provide an error feedback mechanism for the reconstruction of medical videos.
 \citet{lin2020efficient} proposed a network to super-resolve the cardiac MRI slices, which uses bidirectional ConvLSTM as the network backbone. It utilizes domain knowledge of cardiac, and iteratively to enhance low-resolution MRI slices.

\subsection{Surveillance Videos}
\citet{shamsolmoali2019deep} proposed a deep CNN to upsample the low-resolution surveillance videos. The CNN is composed of less than 20 layers and is trained and tested on two surveillance datasets, which are mainly indoor videos.  \citet{lee2018accurate} utilized SRGAN \citep{SRGAN} to enhance the details of the characters on license plate, and they also collected a video dataset with low-resolution and evaluated their method to verify its effectiveness. \citet{guo2020towards} adopted DeblurGAN \citep{kupyn2018deblurgan} to remove the motion blur of the adjacent frames, and then the MEMC was performed on adjacent frames. Finally, high-resolution video frames can be reconstructed through a multi-frame super-resolution algorithm.

In order to super-resolve multi-view face video, \citet{2017Fractional} proposed a fractional-Grey Wolf optimizer-based kernel for the neighboring pixel estimation in the face video. \citet{VFSR} proposed a simple but effective motion-adaptive feedback cell that can capture the motion information and feed it back to the network in an adaptive way for video face super-resolution.

\subsection{3D Reconstruction}
By using the input video sequences, \citet{textureSR} presented a SR method, which produces a 3D mesh of the observed scene with enhanced texture. For multiview video SR methods, \citet{MVSR} adopted the kernel regression to upgrade the information extraction layer, and utilize nonlocal means to information merging layer.  Furthermore,  \citet{3DAPPSR} proposed the first framework that super-resolves the appearance of 3D objects that are captured from multiple view points. The framework combines the power of 2D deep learning-based techniques with the 3D geometric information in the multi-view setting.

\subsection{Virtual Reality}
\citet{liu2020single} proposed a single frame and multi-frame joint super-resolution network, which includes a loss function with weighted mean squared error for the SR of 360-degree panorama videos. They also provided a new panorama video dataset: the MiG Panorama Video for evaluating the panorama VSR algorithms. \citet{9155477} presented a video streaming system to reduce bandwidth requirements for 360-degree videos. The client runs a deep learning based SR model to recover the video, which is heavily compressed at the server. The authors also compared with other state-of-the-art video streaming systems on video quality of experience.


\subsection{Thermal Videos}
 In \citep{app9204405}, a super-resolution model based on  CNN and residual connection was proposed to enhance the thermal videos acquired by thermal cameras, and contactlessly estimate the respiratory rate. Compared with the previous methods, the performance is improved by using super-resolved sequences. \citet{9214230} discussed the performance of SR techniques by using different deep neural networks on benchmark thermal datasets, including SRCNN \citep{SRCNN}, EDSR \citep{EDSR}, autoencoder and SRGAN \citep{SRGAN}. Based on the experimental results, they concluded that SRGAN yields superior performance on thermal frames when comparing with others.

\section{Trends and Challenges}
Although great progress has been made by state-of-the-art video super-resolution methods based on deep learning especially on some public benchmark datasets, there are still challenges and trends discussed below.

\subsection{Lightweight Super-Resolution Models}
The deep learning based video super-resolution methods enjoy high performance, nevertheless they have difficulty in deploying efficiently in many real-world problems. It is noted that their models usually have a mass of parameters and require vast computational and storage resources, and their training also takes a long time. With the popularity of mobile devices in modern life, one expects to apply these models on such devices. To address this issue, several lightweight super-resolution methods have been proposed, e.g., RISTN~\citep{RISTN}, TDAN~\citep{TDAN}, and~\citep{xiao2021space}. How to design and implement a lightweight super-resolution algorithm with high performance for real-world applicants is a major challenge.

\subsection{Interpretability of Models}
Deep neural networks are usually considered as black boxes. That is, we do not know what real information the model learns when the performance is good or bad. In existing video super-resolution models, there is not a theoretical interpretation about how convolution neural networks recover low-resolution video sequences. With a deeper investigation on its interpretation, the performance of super-resolution algorithms for both videos and images may be improved greatly. Some works have paid attention to this problem, e.g.,~\citep{chan2021understanding} and~\citep{liu2021exploit}.

\subsection{Super-Resolution with Larger Scaling Factors}
For video super-resolution tasks, existing works mainly focus on the case of the magnification factors $\times2$, $\times3$ and $\times4$. The more challenging scales such as $\times$8 and $\times$16 have been rarely explored. With the popularity of high-resolution (e.g., 8K and 16K) display devices, larger scaling factors are to be further studied. Obviously, as the scale becomes larger, it is more challenging to predict and restore unknown information in video sequences. This may result in performance degradation for the algorithms, and weaken robustness in the models. Therefore, how to develop stable deep learning algorithms for VSR tasks with larger scaling factors is still challenging. Until now, there is seldom such work on VSR, while several works such as~\citep{chan2021glean} and~\citep{chen2021learning} were proposed for the single image super-resolution tank with larger scaling factors, e.g., $\times$8.

\subsection{Super-Resolution with Arbitrary Scaling Factors}
From Table~\ref{tab:Margin_settings_2}, we can see that most video super-resolution methods are designed for the case of the scaling factor $\times 4$, which is not appropriate for real scenes. On the one hand, other scales like $\times 2$, $\times 3$ or $\times 1.5$ are also very common in VSR tasks. On the other hand, a video super-resolution model with fixed scale will seriously limit its generalization and portability. Therefore, a universal VSR method for arbitrary scale factors is greatly needed in real-world applications. Several works about image super-resolution with arbitrary scaling factors have been presented, e.g.,~\citep{meta-sr2019}, and~\citep{wang2021learning}, while the works on arbitrary scale factor upsampling for videos are still seldom.

\subsection{More Reasonable $\&$ Proper Degradation Process of Videos}
In existing works, the degraded LR videos are attained through the two methods: One is directly downsampling HR videos by using interpolation, such as bicubic. The other is performing the Gaussian blurring on HR videos and then downsampling the video sequences. Although both methods perform well in theory, they always perform poorly in practice. As it is known, the real-world degradation process is very complex and includes much uncertainty. The blurring and interpolation are not adequate for modeling this problem. Therefore, when constructing LR videos, the degradation should be modeled theoretically in consistent with the real-world case to reduce the gap between research and practice. There are a few works involving the degradation process of videos for super-resolution, such as~\citep{zhang2018spatio}.

\subsection{Unsupervised Super-Resolution Methods}
Most state-of-the-art VSR methods adopt a supervised learning paradigm. In other words, the deep neural networks require a large number of paired LR and HR video frames for training. However, such paired datasets are hard or costly to obtain in practice. One may synthesize the LR/HR video frames, the performance of super-resolution methods is still not satisfied as the degradation model is too simple to characterize the real-world problem and results in inaccurate HR/LR datasets. Thus, unsupervised VSR methods are highly demanded. Some works of unsupervised VSR on satellite videos have been proposed, e.g.,~\citep{he2020unsupervised}, but not about generic videos.

\subsection{More Effective Scene Change Algorithms}
Existing video super-resolution methods rarely involve the videos with scene change. In practice, a video sequence usually has many different scenes. When we consider the problem of video super-resolution on such videos, they have to be split into multiple segments without scene change and processed individually. This may result in large computational time. In fact, a simple subnet in 3DSRnet~\citep{3DSRnet} has been proposed to deal with scene change, and it includes scene boundary detection and frame replacement. More dedicated networks that can process videos with complex scene changes are necessary for real-world applications.

\subsection{More Reasonable Evaluation Criteria for Video Quality}
The criteria for evaluating the quality of super-resolution results mainly include PSNR and SSIM. However, their values are not able to reflect the video quality for human perception. That is, even if the PSNR value of a recovered video is high, the video also makes people uncomfortable. Therefore, new evaluation criteria for videos that are consistent with human perception need to be developed. More attentions have been attracted to the quality evaluation for images, such as~\citep{gu2020image}. However, the video quality including coherence between frames will be investigated in the future.

\subsection{More Effective Methods for Leveraging Information}
An important characteristic of video super-resolution methods is leveraging the information contained in video frames. The effectiveness of utilization influences the performance directly. Although many methods have been proposed, as mentioned in this paper, there are still some disadvantages. For instance, 3D convolution and non-local modules require a large amount of computation, and the accuracy of optical estimation can not be guaranteed. Therefore, the methods that can effectively utilize information contained in different frames is worth further studying.

\section{Conclusions}
In this survey, we reviewed the development of deep learning approaches for video super-resolution in recent years. We first classified existing video super-resolution algorithms into seven subcategories by the way of leveraging information contained in video frames, described the key ideas of representative methods and summarized the advantages and disadvantages of each method. Furthermore, we also compared and analyzed the performance of those methods on benchmark datasets, and outlines the wide applications of video super-resolution algorithms. Although the deep learning based VSR methods have made great progress, we listed eight open issues for the development of VSR algorithms, which is expected to provide some enlightment for researchers.


\section*{Acknowledgment}
We thank all the reviewers for their valuable comments. We would like to thank Mr.\ \textbf{Zekun Li} (Master student at School of Artificial Intelligence in Xidian University) and Dr.\ \textbf{Yaowei Wang} (Associate Professor with Peng Cheng Laboratory, Shenzhen, China) for their help in improving the quality of this manuscript. This work was supported by the National Natural Science Foundation of China (Nos.\ 61976164, 61876220, 61876221, and 61906184).

\section*{Author Biography}
\textbf{Hongying Liu} received her B.E. and M.Sc. degrees in Computer Science and Technology from Xi'An University of Technology, China, in 2006 and 2009, respectively, and Ph.D. in Engineering from Waseda University, Japan in 2012. Currently, she is a faculty member at the School of Artificial Intelligence, and also with the Key Laboratory of Intelligent Perception and Image Understanding of Ministry of Education, Xidian University, China. In addition, she is a senior member of IEEE. Her major research interests include image and video processing, intelligent signal processing, machine learning, etc.

\textbf{Zhubo Ruan} received his M. Sc. degree from School of Artificial Intelligence in Xidian University in 2021. His research interests include machine learning and video super-resolution, etc.

\textbf{Peng Zhao} is currently pursuing the M.Sc. degree with the School of Artificial Intelligence in Xidian University. His research interests include video super-resolution, medical image processing, etc.

\textbf{Chao Dong} received the Ph.D.\ degree from The Chinese University of Hong Kong in 2016, advised by Prof.\ Tang and Prof.\ Loy. He is currently an Associate Professor with the Shenzhen Institute of Advanced Technology, Chinese Academy of Sciences. His current research interests include low-level vision problems, such as image/video super-resolution, denoising, and enhancement. His team won the first place in international super-resolution challenges: NTIRE2018, PIRM2018, and NTIRE2019.

\textbf{Fanhua Shang} received the Ph.D. degree in Circuits and Systems from Xidian University, Xi'an, China, in 2012. He is currently a professor with the School of Artificial Intelligence, Xidian University, China. Prior to joining Xidian University, he was a Research Associate with the Department of Computer Science and Engineering, The Chinese University of Hong Kong. From 2013 to 2015, he was a Post-Doctoral Research Fellow with the Department of Computer Science and Engineering, The Chinese University of Hong Kong. From 2012 to 2013, he was a Post-Doctoral Research Associate with the Department of Electrical and Computer Engineering, Duke University, Durham, NC, USA. His current research interests include machine learning, data mining, and computer vision.

\textbf{Yuanyuan Liu} received the Ph.D. degree in Circuits and Systems from Xidian University, Xi'an, China, in 2012. He is currently a professor with the School of Artificial Intelligence, Xidian University, China. Prior to joining Xidian University, he was a Research Associate with the Department of Computer Science and Engineering, The Chinese University of Hong Kong. From 2013 to 2015, he was a Post-Doctoral Research Fellow with the Department of Computer Science and Engineering, The Chinese University of Hong Kong. From 2012 to 2013, he was a Post-Doctoral Research Associate with the Department of Electrical and Computer Engineering, Duke University, Durham, NC, USA. His current research interests include machine learning, data mining, and pattern recognition.

\textbf{Linlin Yang} received the M.Sc. degree in circuits and systems from Taiyuan University of Science and Technology, Taiyuan, China, in 2020. She is currently pursuing the Ph.D. degree with the School of Artificial Intelligence in Xidian University, China. Her research interests include image processing, etc.

\textbf{Radu Timofte} is a professor and chair for computer vision at University of Wurzburg, Germany and the recipient of a 2022 Humboldt Professorship Award for Artificial Intelligence. He is also a lecturer and group leader at ETH Zurich, Switzerland. He obtained his PhD degree in Electrical Engineering at the KU Leuven, Belgium in 2013, the MSc at the Univ. of Eastern Finland in 2007, and the Dipl. Eng. at the Technical Univ. of Iasi, Romania in 2006. He is co-founder of Merantix, co-organizer of NTIRE, CLIC, AIM, MAI and PIRM events, and member of IEEE, CVF, and ELLIS.  He is associate/area editor for journals such as IEEE Trans. PAMI, Elsevier Neurocomputing, Elsevier CVIU and SIAM Journal on Imaging Sciences and he regularly serves as area chair/SPC for conferences such as CVPR, ICCV, ECCV, IJCAI. His current research interests include deep learning, visual tracking, mobile AI, image/video compression, restoration, manipulation and enhancement.


\bibliographystyle{spbasic}		
\bibliography{refff}
		
\end{document}